\documentclass[sigconf]{aamas} 

\usepackage{balance} 
\usepackage{amsmath}

\usepackage[capitalise]{cleveref}
\usepackage{algorithm, algpseudocode}

\usepackage{fp,tikz,pgfplots}

\usepackage{tabularx}
\usepackage{caption}
\usepackage{subcaption}
\usepackage{bm}
\usepackage{lipsum}

\newtheorem{assumption}{Assumption}
\newtheorem{lemma}{Lemma}
\newtheorem{theorem}{Theorem}

\newtheorem{remark}{Remark}

\xdefinecolor{blueZ}{RGB}{0,77,163}
\xdefinecolor{redZ}{RGB}{232,61,117}

\setcopyright{ifaamas}
\acmConference[AAMAS '24]{$\text{Preprint.}$ Proc.\@ of the 23rd International Conference
on Autonomous Agents and Multiagent Systems (AAMAS 2024)}{May 6 -- 10, 2024}
{Auckland, New Zealand}{N.~Alechina, V.~Dignum, M.~Dastani, J.S.~Sichman (eds.)}
\copyrightyear{2024}
\acmYear{2024}
\acmDOI{}
\acmPrice{}
\acmISBN{}

\acmSubmissionID{434}

\title[Elective Learning for Distributed GPR]{Whom to Trust? \\Elective Learning for Distributed Gaussian Process Regression}

\author{Zewen Yang$^*$}
\affiliation{
  \institution{Robert Koch Institute}
  \city{Berlin}
  \country{Germany}}
\email{yangz@rki.de}

\author{Xiaobing Dai$^*$}
\affiliation{
  \institution{Technical University of Munich}
  \city{Munich}
  \country{Germany}}
\email{xiaobing.dai@tum.de}

\author{Akshat Dubey}
\affiliation{
  \institution{Robert Koch Institute}
  \city{Berlin}
  \country{Germany}}
\email{dubeya@rki.de}

\author{Sandra Hirche}
\affiliation{
  \institution{Technical University of Munich}
  \city{Munich}
  \country{Germany}}
\email{hirche@tum.de}

\author{Georges Hattab}
\affiliation{
  \institution{Robert Koch Institute\\
  Freie Universität Berlin}
  \city{Berlin}
  \country{Germany}}
\email{hattabg@rki.de}

\begin{abstract}
This paper introduces an innovative approach to enhance distributed cooperative learning using Gaussian process (GP) regression in multi-agent systems (MASs). The key contribution of this work is the development of an elective learning algorithm, namely prior-aware elective distributed GP (Pri-GP), which empowers agents with the capability to selectively request predictions from neighboring agents based on their trustworthiness. The proposed Pri-GP effectively improves individual prediction accuracy, especially in cases where the prior knowledge of an agent is incorrect. Moreover, it eliminates the need for computationally intensive variance calculations for determining aggregation weights in distributed GP. Furthermore, we establish a prediction error bound within the Pri-GP framework, ensuring the reliability of predictions, which is regarded as a crucial property in safety-critical MAS applications.
\end{abstract}


\newcommand{\BibTeX}{\rm B\kern-.05em{\sc i\kern-.025em b}\kern-.08em\TeX}

\begin{document}


\pagestyle{fancy}
\fancyhead{}


\maketitle 

\renewcommand{\thefootnote}{\fnsymbol{footnote}}
\footnotetext[1]{Equal contribution.}
\renewcommand{\thefootnote}{\arabic{footnote}}

\section{Introduction}
In the context of multi-agent systems (MASs), distributed learning entails a collaborative approach where one or more groups of agents join forces to improve their understanding and knowledge of complex tasks, such as robotic swarm navigation \cite{6224987}, underwater vehicle resource exploration missions \cite{8867291,yanVirtualLeaderBased2020}, and air drone search and rescue operations \cite{8695011}, etc. To address the inherent challenges posed by uncertain dynamics or environmental conditions in dynamic systems, distributed learning integrates supervised machine learning techniques enabling agents to learn cooperatively. This approach leads to more effective and robust learning capabilities compared to traditional single-agent models \cite{provost1996scaling}.

Specifically, Neural Networks (NNs) emerge as the prevailing methodology approximating complex mappings or functions in MASs \cite{daiEventTriggeredDistributedCooperative2019}. To learn the unknown patterns jointly, the NN weights are shared among neighboring agents, facilitating the attainment of optimal parameter values \cite{wangCooperativeLearningNeural2017, gaoNeuralNetworkBasedDistributed2020}. Several research endeavors have been dedicated to system identification within the framework of addressing uncertainties~\cite{jafariIntelligentControlUnmanned2018, daiDistributedCooperativeLearning2021}. 
However, as the complexity of NN models increases with additional hidden layers and neurons, the practicality of sharing all NN weights within constrained communication bandwidth becomes unfeasible. 
This challenge further leads to significant delays in the learning process, rendering it impractical and resulting in the MAS dynamics potentially unstable. 
Though only exchanging the predictions, the inherent challenge arises from the inadequacy in precisely quantifying prediction uncertainties, thereby impeding its applicability in safety-critical tasks.

An alternative supervised machine learning approach, Gaussian process regression (GPR) \cite{rasmussenGaussianProcessesMachine2006}, has been widely used in the realm of safety-critical control systems, primarily owing to its distinctive attributes. Under the Bayesian inference framework, GPR not only provides probabilistic predictions, where the prior model can be updated continuously accommodating the incorporation of observations \cite{schurchRecursiveEstimationSparse2020}, but also offers error bounds endowed with robust guarantees~\cite{ledererUniformErrorBounds2019}.  In contrast to NNs-based methods, GP models are only required to share their individual predictions with connected counterparts~\cite{yin2023learningbased}. Even if the agent lacks access to the complete training dataset, collaborative improvements in prediction quality can be achieved by aggregating predictions from neighboring agents. The synergy of aggregated predictions from neighboring agents, as detailed in prior research \cite{yangDistributedLearningConsensus2021,ledererCooperativeControlUncertain2023,yangCooperativeLearning2024}, underscores the effectiveness of this approach in achieving improved prediction quality within the multi-agent framework. Moreover, to improve the efficiency of cooperative learning with GPs, the event-triggered mechanism is introduced in \cite{dai2024cooperative,dai2024DecentralizedEventTriggered}. However, the mentioned literature above based on distributed GP \cite{trespMixturesGaussianProcesses2000, deisenrothDistributedGaussianProcesses2015} imposes a constraint in the sense that it mandates the exchange of information with all neighboring agents, offering no flexibility for agents to selectively determine which neighbors to collaborate with. The agent needs to aggregate the predictions from the neighbors, thus each agent is compelled to compute predictions for all of its neighboring counterparts, potentially incurring a substantial increase in computational overhead. For example, the product of GP experts (POE) methods \cite{cao2015generalized} and Bayesian committee machine (BCM) methods~\cite{10.1162/089976600300014908, pmlr-v80-liu18a} rely on the posterior variance of GP requiring $\mathcal{O}(N^2)$ calculations. This concern becomes particularly salient in scenarios characterized by limited computational resources at each agent's disposal or when the expeditious generation of predictions is imperative. Furthermore, the rigidity of this collaborative setup raises noteworthy issues, especially in cases where the prior knowledge is erroneous. This is ignored by most distributed GP approaches, for instance, the mixture of GP experts (MOE) approach \cite{trespMixturesGaussianProcesses2000}, where the aggregation weight is just the reciprocal of the total number of the GP models. In such instances, a uniform collaboration approach may yield suboptimal results, as it does not accommodate the possibility that an individual agent may possess superior predictive capabilities compared to its collaborators.

To address these challenges, we propose a novel approach where agents are empowered with the capability of requesting predictions exclusively, allowing them to actively select their collaborators from among their neighbors. This elective learning method leverages the error between prior knowledge and real observations, called prior-aware elective distributed GP (Pri-GP), which can let the agent smartly choose the neighbors who are worth trusting. Therefore, it not only reduces the heavy computation required by the neighbors but also improves the individual agent's prediction in the distributed cooperative learning framework avoiding aggregating the potentially misleading prediction from the neighbors whose prior knowledge is significantly wrong. 

The contribution of this paper is that an elective distributed cooperative learning algorithm for MASs with distributed GPR is proposed. The primary innovation lies in the proposal of an error metric that leverages prior errors, which possesses broader applicability beyond its immediate application and can be seamlessly integrated into various machine learning methodologies. In particular, the proposed Pri-GP approach offers a remarkable degree of flexibility by circumventing the necessity of computing variance for the determination of aggregation weights, a process commonly associated with a computational complexity of $\mathcal{O}(N^2)$ in distributed GP techniques. Additionally, we provide a prediction error bound using the Pri-GP framework, thereby ensuring the reliability of predictions, a crucial aspect, particularly in the context of safety-critical applications.


\section{Preliminaries and Problem Setting}

\subsection{Notation and Graph Theory}\label{subsec_notation}
We use the notation $\mathbb{R}_{+} /  \mathbb{R}_{0,+}$ to denote positive real numbers without/with zero and denote natural numbers without/with zero as $\mathbb{N} / \mathbb{N}_{0}$, respectively. Unless otherwise specified, the identity matrix is $\boldsymbol{I}$, and a matrix or vector consisting of elements 1 is denoted by $\mathbf{1}$, with appropriate sizes as needed. The Euclidean norm of a vector or matrix is represented as $\|\cdot\|$, the cardinality of a set $\mathcal{N}$ as $\left | \mathcal{N} \right |$, and the element-wise absolute operator for vector input $\boldsymbol{v}$ as $|\boldsymbol{v}|$.

To characterize the communication network of the distributed MAS, an undirected graph $\mathcal{G} = (\mathcal{V}, \mathcal{E} )$ is employed among $S$ agents, $S \in \mathbb{N}$. 
The node set $\mathcal{V} = \{1, \ldots, S\}$ represents the index of the agents, and $\mathcal{E} \subseteq \mathcal{V} \times \mathcal{V}$ signifies the set of edges between nodes, where an edge $\left ( i, j \right ) \in \mathcal{E}$ indicates that agent $i$ and agent $j$ exchange their information between each other. The self-loop included adjacency matrix of $\mathcal{G}$ is denoted by $\mathcal{A} = \left [a_{ij} \right ]\in  \mathbb{R}^{S\times S}$, where all diagonal entries $a_{ii} = 1$, other elements of the matrix $a_{ij} = a_{ji} = 1$ if $\left (i,j \right )  \in \mathcal{E} $ and $ a_{ij}=0 $ otherwise. Moreover, we let the set of neighbours of agent $i$ represent as $\mathcal{N}_i= \{j\in \mathcal{V}:(i,j)\in {\mathcal{E}}\}$ and the set $\bar{\mathcal{N}}_i$ comprises agent $i$ along with its neighbors, meaning it contains agent $i$ itself and all the agents in $\mathcal{N}_i$.

\subsection{Problem Description}\label{subsec_problem}
In this work, we delve into the investigation of a distributed MAS comprising a total of $S$ individual agents. The primary objective of these homogeneous agents is to identify the identical functional characteristics of their dynamical systems
\begin{align}
   \dot{\boldsymbol{x}} = \boldsymbol{f}(\boldsymbol{x}),
\end{align}
considering $\boldsymbol{f}(\boldsymbol{x}) = [f_1(\cdot), \cdots, f_m(\cdot)]^{\rm T}$, where for each dimension, $f_d(\cdot): \mathbb{X}  \to \mathbb{R}$, $\forall d = 1, \cdots, m$ and $\mathbb{X}\subset \mathbb{R}^{m}$ is a $m\in\mathbb{N}$ dimensional compact domain. These agents actively engage in communication with one another through a network, herein referred to as $\mathcal{G}$. Concurrently, each agent has the collection of observational data set $\mathbb{D}_i$ with the subscript $i$ indicating the specific agent within the set $\mathcal{V}$, which are subsequently harnessed for the purpose of estimating the unknown functions. It is noteworthy that each agent possesses its own distinct set of prior knowledge pertaining to these unidentified functions, which is represented as $\hat{f}(\cdot)$ considering the same mapping relationship of $f$. To facilitate the utilization of this prior knowledge and the learned functions for describing the system's dynamics, we introduce the following assumption.

\begin{assumption}\label{ass_unknown_f}
The functions $f(\cdot)$ and $\hat{f}(\cdot)$ exhibit local Lipschitz continuity within the compact domain $\mathbb{X}$, characterized by a Lipschitz constant denoted as $L_f \in \mathbb{R}$, i.e, $\|\nabla {f}(\boldsymbol{x})\| \le L_{{f}}$  for all $\boldsymbol{x} \in \mathbb{X}$.
\end{assumption}
This assumption is frequently encountered in the context of nonlinear systems \cite{khalil2015nonlinear}, which serves the guarantee of the existence and uniqueness of solutions for nonlinear autonomous systems. In practice, this assumption merely necessitates the system's continuity, with the subsequent establishment of Lipschitz continuity being a derived property within the bounded region denoted as $\mathbb{X}$. Consequently, it can be contended that this assumption imposes no onerous constraints on the system under consideration.

It is important to highlight that in this paper, we have chosen to work with a one-dimensional estimation, i.e., a scalar function, where $d = 1$. This simplification in dimensionality has been adopted for the sake of keeping our notations concise and straightforward. However, it should be noted that the outcomes derived in this work can be readily extended to higher dimensional functions achieved by employing techniques such as the Kronecker product and multi-output learning methods.

To describe the training data set of the agent $i$ comprising streaming data pairs, we denote it as $\mathbb{D}_i$ with $N_i\in \mathbb{N}$ training data pairs. This data set is represented as $\mathbb{D}_i=\big\{\big( \boldsymbol{x}_i^{(p)}, y_i^{(p)} \big)\big\}_{p=1,\dots, N_i}$, where each pair consists of a training input $\boldsymbol{x}_i \in \mathbb{X}$ and a corresponding training output $y_i\in \mathbb{R}$, and satisfies the following assumption.
\begin{assumption} \label{ass_dataset}
    The data pair $\big\{ \big(\boldsymbol{x}_i^{(p)}, y_i^{(p)} \big)\big\}$ that is obtained by agent $i \in \mathcal{V}$, such that the noise $\varsigma_i^{(p)} \in \mathbb{R}$ of $y_i^{(p)} = f\big(\boldsymbol{x}_i^{(p)}\big) + \varsigma_i^{(p)}$ follows a zero-mean, independent and identical Gaussian distribution, i.e., $\varsigma_i^{(p)} \sim \mathcal{N}(0, \sigma_{n}^2)$, $\forall p \in \mathbb{N}$ with $\sigma_{n} > 0$.
\end{assumption}
As outlined in \cref{ass_dataset}, it is assumed that each agent independently collects their own dataset without sharing it among others. While this assumption necessitates precise and complete measurements of the system states, a requirement commonly encountered in MASs when employing data-driven methodologies, it is possible to effectively address the measurement noise associated with the variable $\boldsymbol{x}$. This can be achieved through various techniques, such as Taylor expansion, as demonstrated in prior works like \cite{mchutchonGaussianProcessTraining2011,kimModelReferenceGaussian2023}, or by incorporating noise handling directly into the kernel function, as discussed in \cite{wang2022gaussian}. It is worth noting that there are broader considerations related to noise distribution relaxation, we refer to \cite{chowdhuryKernelizedMultiarmedBandits2017, maddalenaDeterministicErrorBounds2021}. However, these aspects lie beyond the scope of this paper.

In the development of a distributed learning framework for the MAS, the estimation of the unknown function $f(\cdot)$ at time $t_k \in \mathbb{R}_{0,+}$ of the $i$-th agent is considered as
\begin{equation}\label{eq_tilde_f}
    \tilde{f}_i(\boldsymbol{x}(t_k)) = \sum_{j=1}^S \omega_{ij}\phi_{ij}\Big(\boldsymbol{x}(t_k), \mathbb{D}_j, \hat{f}_j(\boldsymbol{x}(t_k))\Big), ~ i \in \mathcal{V},
\end{equation}
where $k\in \mathbb{N}$ is indicating the specific instance of prediction, and the cooperative estimation function $\phi_{ij}(\cdot, \cdot, \cdot )$ corresponds to the prediction mechanism employed by each agent. Specifically, when $i \neq j$, it characterizes the prediction generated through the neighbor agent $j$ with its training data set $\mathbb{D}_j$ and prior knowledge function $\hat{f}_j(\cdot)$. While $i=j$, it represents the prediction of agent $i$ itself using its own prior and training data set. 

Therefore, the primary focus of this paper revolves around the development of a collaborative estimation function tailored to augment individual predictions of an unknown function in an elective manner, which can enhance the predictions without necessitating the aggregation of predictions from all neighboring agents, thereby mitigating the computational burden imposed on these neighbors. Importantly, this elective strategy takes into account the varying accuracy of prior knowledge possessed by agents, which is illustrated in \cref{sec_ActiveCooperativeLearning}.

\subsection{Gaussian Process Regression}\label{subsec_GPR}
In this paper, Gaussian process regression, a supervised machine learning technique, is employed to perform inference on the unknown function $f(\cdot)$. A Gaussian process $\mathcal{GP}( \hat{f}(\cdot), \kappa(\cdot,\cdot) )$ is utilized to establish a probabilistic model characterized by two fundamental components: the prior mean function $\hat{f}(\cdot)$, and the kernel function, denoted as $\kappa(\cdot, \cdot): \mathbb{X} \times \mathbb{X} \rightarrow \mathbb{R}_{0, +}$, which satisfies the following assumption.
\begin{assumption}\label{ass_kernel}
    The kernel function $\kappa (\|{x} - x'\|) = \kappa (x,x')$  is chosen as stationary, monotonically decreasing, and Lipschitz continuous with a specified Lipschitz constant denoted as $L_{\kappa}$.
\end{assumption}
The adoption of a Lipschitz continuous kernel emerges as a judicious selection when dealing with continuous unknown functions within a confined domain. The kernel's monotonic decrease, as reflected in its behavior, implies a diminishing strength of association between the training data and the evaluated point as their Euclidean distance increases. Therefore, a common choice of kernel function is ARD exponential kernel formulated as \looseness=-1
\begin{align}
    \kappa\left(\boldsymbol{x}, \boldsymbol{x}^{\prime}\right)=\sigma_{r}^{2} \exp \Big(-\frac{1}{2} \sum_{j=1}^{m} {l_{j}}^2(x_{j}-x_{j}^{\prime})^2\Big),
\end{align}
where $\boldsymbol{x} = [x_1, x_2, \dots, x_m] \in \mathbb{X}$.  $\sigma_r\in\mathbb{R}_+$ and $l_j\in\mathbb{R}_+$ are hyper-parameters.  

We consider the agent $i\in \mathcal{V}$ within the MAS to be equipped with a GP model characterized by hyperparameters denoted as $\boldsymbol{\theta}=\{\sigma_{r}, l_{s}, s= 1,2,\!\dots\!,m\}$. Additionally, the agent possesses a fixed training dataset denoted as $\mathbb{D}_i$ with $N_i\in \mathbb{N}_0$ training data pairs under \cref{ass_dataset} and holds different prior of the unknown function $\hat{f}_i(\cdot)$ satisfying \cref{ass_unknown_f}. Agent $i$ performs predictions at discrete time points denoted as $t_{k}$. These predictions are represented as the posterior mean prediction and associated prediction variance \cite{rasmussenGaussianProcessesMachine2006} at the query point $\boldsymbol{x}(t_{k})$, which are formulated as 
\begin{align}
\label{eq_mu} \mu_i\big(\boldsymbol{x}(t_k)|\hat{f}_i(\boldsymbol{x}(t_k)), \mathbb{D}_i\big) =&~ \hat{f}_i(\boldsymbol{x}(t_{k})) \\
&  + {K}\big(\boldsymbol{x}(t_{k}), \boldsymbol{X}_i\big){\boldsymbol{K}}(\boldsymbol{X}_i)^{-1}\big(\boldsymbol{Y}_i^{\rm T} - \hat{\boldsymbol{f}}_i(\boldsymbol{X}_i)^{\rm T}\big),  \nonumber  \\
\label{eq_simga}    \sigma_i\big(\boldsymbol{x}(t_{k})|\hat{f}_i(\boldsymbol{x}(t_k)),\mathbb{D}_i\big) =&~ 
    \kappa(\boldsymbol{x}(t_{k}), \boldsymbol{x}(t_{k})) \\
   & - {K}\big(\boldsymbol{x}(t_{k}), \boldsymbol{X}_i\big){\boldsymbol{K}}(\boldsymbol{X}_i)^{-1}{K}\big(\boldsymbol{x}(t_{k}), \boldsymbol{X}_i\big)^{\rm T}, \nonumber
\end{align}
respectively, where
\begin{align}
\label{eq_Kinv}&{\boldsymbol{K}}(\boldsymbol{X}_i) = \mathcal{K}\big(\boldsymbol{X}_i, \boldsymbol{X}_i\big) + \sigma_n^2\boldsymbol{I},\\
\label{eq_Kmatrix}&\mathcal{K}(\boldsymbol{X}_i,\boldsymbol{X}_i) = \big[ \kappa \big( \boldsymbol{x}_i^{(a)},\boldsymbol{x}_i^{(b)} \big) \big]_{a,b = 1, \dots, N_i}\\
\label{eq_Kcov}&{K}\big(\boldsymbol{x}(t_{k}), \boldsymbol{X}_i(t_{k})) = \big[ \kappa 
\big( \boldsymbol{x}(t_{k}),\boldsymbol{x}_i^{(1)} \big) \cdots \kappa 
\big( \boldsymbol{x}_{t_{k}},\boldsymbol{x}_i^{(N_i)} \big) \big], \\
\label{eq_XY}&\boldsymbol{X}_i = \big[ {{\boldsymbol{x}_i^{(1)}} \cdots {\boldsymbol{x}_i^{(N_i)}}} \big], \boldsymbol{Y}_i = \big[ {{y_i^{(1)}} \cdots {y_i^{(N_i)}}} \big], 
\end{align}
and the concatenated prior mean value is defined as $\hat{\boldsymbol{f}}(\boldsymbol{X}_i) = \big[\hat{{f}}\big(\boldsymbol{x}_i^{(1)}\big), \dots, \hat{{f}}\big(\boldsymbol{x}_i^{(N_i)}\big)\big]$. 

Therefore, each agent $i$ can use the posterior mean to identify the unknown $f(\cdot)$ at $\boldsymbol{x}(t_k)$ by \cref{eq_mu}, the individual estimation function $\phi_{ii}\big(\boldsymbol{x}(t_k), \mathbb{D}_i, \hat{f}_i(\boldsymbol{x}(t_k))\big) = \mu_i\big(\boldsymbol{x}(t_k)|\hat{f}(\boldsymbol{x}(t_k)), \mathbb{D}_i\big) $. To simplify our notation, we denote $\mu_i(\boldsymbol{x}(t_k)|\hat{f}_i(\boldsymbol{x}(t_k)), \mathbb{D}_i)$ as $\mu_i(\boldsymbol{x}(t_k))$ in the subsequent sections of this paper. Having established the GPR as our foundational tool, the subsequent section is dedicated to the formulation of our elective distributed learning approach.

\section{Elective Distributed Learning with Prior-Aware GPR}\label{sec_ActiveCooperativeLearning}
To assess the reliability of the collaborators, we introduce the prior estimation error in \cref{subsec_priorError}. This metric serves as a quantitative measure for gauging the trustworthiness of neighboring agents in the MAS. Building upon this quantitative foundation, we proceed to formulate an elective distributed learning algorithm, as outlined in \cref{subsec_PriorAwareLearning}. Additionally, to bolster the safety and guarantee of the learning scenario, we establish a prediction error bound within the Pri-GP framework in \cref{subsec_predictionErrorBound}.

\subsection{Prior Error Quantification}\label{subsec_priorError}
In multi-agent systems, each agent possesses a finite training dataset, which naturally leads to a scenario where predictions for points lying beyond the training data domain or in sparsely sampled regions become highly reliant on prior knowledge. Consequently, the accuracy of these predictions is predominantly influenced by the quality of the prior information in the Bayesian framework. This situation underscores the potential challenges arising from incorrect or inadequate prior knowledge. Therefore, there arises a compelling need for a systematic mechanism to assess the degree of inaccuracy associated with prior knowledge, particularly in the presence of observed true values for predictions. To formally quantify the disparity between the prior estimation and observed data, we introduce the concept of prior estimation error denoted by 
\begin{equation} \label{eq_prior_estimation_error}
    e_i(\boldsymbol{x}(t_{k})) = \hat{f}_i(\boldsymbol{x}(t_{k})) - y(t_{k}).
\end{equation}
Given the availability of system observations, we systematically log the associated errors. To establish the evolving cumulative error over time, we define the average accumulated historical prior estimation error as
\begin{equation}\label{eq_varepsilon}
    \varepsilon_{i}(t_{k})= \frac{1}{k}{ \sum_{l=1}^{k} |e_i(\boldsymbol{x}(t_{k-l}))|}.
\end{equation}
This metric serves as a pivotal instrument in characterizing the deviation between prior expectations and empirical observations, thereby enhancing our capacity to evaluate and interpret the reliability of the models, which is illustrated in the following lemma.
\begin{lemma} \label{lemma_varepsilon}
    The variable $\varepsilon_{i}(t_{k})$ reflects the prediction error on the training data set $\mathbb{D}_i$, and the measurement error.
    In particular, $\varepsilon_{i}(t_{k})$ is written as
    \begin{align} \label{eq_varepsilon_xi_varsigma}
        \varepsilon_{i}(t_{k}) =  \frac{1}{k} \mathbf{1}^{\rm T} | \boldsymbol{G}_k (\boldsymbol{\xi}_k + \boldsymbol{\varsigma}_k ) |,
    \end{align}
    where $\boldsymbol{G}_k =  (\boldsymbol{I} + \sigma_{n}^{-2} \mathcal{K}(\boldsymbol{X}_i, \boldsymbol{X}_i) )$. The aggregated error denotes $\boldsymbol{\xi}_k = [\xi_0, \cdots, \xi_k]^T$ with $\xi_p = f(\boldsymbol{x}^{(p)}) - \mu_i(\boldsymbol{x}^{(p)}), \forall p = 1,\cdots, k$. The noise vector $\boldsymbol{\varsigma}_k$ collects all measurement from $t = t_0$ to $t_k$, i.e., $\varsigma_k = [ \varsigma^{(0)}, \cdots, \varsigma^{(k)}]^T$, where the individual noise $\varsigma^{(p)}, \forall p = 1,\cdots, k$, follows \cref{ass_dataset}.
\end{lemma}
\begin{proof}
    Considering the prediction of each sample in $\mathbb{D}_i$ using \eqref{eq_mu}, the aggregated prediction $\boldsymbol{\mu}_i(\boldsymbol{X}_i) = [\mu_i(\boldsymbol{x}^{(0)}), \cdots, \mu_i(\boldsymbol{x}^{(N_i)})]^T$ is written as
    \begin{align}
        \boldsymbol{\mu}_i(\boldsymbol{X}_i) =&\hat{\boldsymbol{f}}_i(\boldsymbol{X}_k)^{\rm T} + \mathcal{K}(\boldsymbol{X}_i, \boldsymbol{X}_i) \boldsymbol{K}(\boldsymbol{X}_i)^{-1} \big(\boldsymbol{Y}_k^{\rm T} - \hat{\boldsymbol{f}}_i(\boldsymbol{X}_i)^{\rm T}\big) \\
        =& \big( \boldsymbol{I} - \mathcal{K}(\boldsymbol{X}_i, \boldsymbol{X}_i) \boldsymbol{K}(\boldsymbol{X}_i)^{-1} \big) \hat{\boldsymbol{f}}_i(\boldsymbol{X}_i)^{\rm T} \nonumber \\
        &+ \mathcal{K}(\boldsymbol{X}_i, \boldsymbol{X}_i) \boldsymbol{K}(\boldsymbol{X}_i)^{-1} \boldsymbol{Y}_k^{\rm T}. \nonumber
    \end{align}
    Then, the aggregated deviation between the measurements $y^{(p)}$ and the prediction $\mu_i(\boldsymbol{x}^{(p)})$ denoted by
    \begin{align}\label{eq_Y_mu}
        \boldsymbol{Y}_k^{\rm T} - \boldsymbol{\mu}_i(\boldsymbol{X}_k) &= \sigma_{n}^{2} \boldsymbol{K}(\boldsymbol{X}_i)^{-1} \big(\boldsymbol{Y}_k^{\rm T} - \hat{\boldsymbol{f}}_i(\boldsymbol{X}_k)^{\rm T}\big) \nonumber \\
        &= \boldsymbol{G}_k^{-1} \big(\boldsymbol{Y}_k^{\rm T} - \hat{\boldsymbol{f}}_i(\boldsymbol{X}_k)^{\rm T}\big),
    \end{align}
    which is also equivalent to 
    $$\boldsymbol{Y}_k^{\rm T} - \hat{\boldsymbol{f}}_i(\boldsymbol{X}_k)^{\rm T} = \boldsymbol{G}_k \big(\boldsymbol{Y}_k^{\rm T} - \boldsymbol{\mu}_i(\boldsymbol{X}_k)\big)$$ 
    due to the non-singular $\boldsymbol{G}_k$.
    Note that $\boldsymbol{Y}_k^{\rm T} - \boldsymbol{\mu}_i(\boldsymbol{X}_k)$ is divided into two parts with prediction error $\boldsymbol{\xi}_k$ and measurement error $\varsigma_k$, i.e., $\boldsymbol{Y}_k^{\rm T} - \boldsymbol{\mu}_i(\boldsymbol{X}_k) = \boldsymbol{\xi}_k + \varsigma_k$.
    Moreover, reformulating \eqref{eq_varepsilon} as $\varepsilon_{i}(t_{k}) = {k}^{-1} \boldsymbol{1}^{\rm T} \boldsymbol{e}_i(\boldsymbol{X}_k)$ with $\boldsymbol{e}_i(\boldsymbol{X}_k) = [e_i(\boldsymbol{x}(t_0)), \cdots, e_i(\boldsymbol{x}(t_k))]^T$, the accumulated prior estimation error $\varepsilon_{i}(t_{k})$ is written as
    \begin{align}
        \varepsilon_{i}(t_{k}) = \frac{1}{k} \boldsymbol{1}^{\rm T} |\boldsymbol{Y}_k^{\rm T} - \boldsymbol{\mu}_i(\boldsymbol{X}_k)|
    \end{align}
    and then the result in \eqref{eq_varepsilon_xi_varsigma} is derived.
\end{proof}
\begin{remark}
    \cref{lemma_varepsilon} shows the accumulated historical prior estimation error encodes the joint effects of the prediction and measurement. Moreover, the coefficient matrix $\boldsymbol{G}_k$ indicates the correlation of the training data by $\mathcal{K}(\boldsymbol{X}_i, \boldsymbol{X}_i)$.
    As the value of $\boldsymbol{G}_k$ is influenced by the number of training samples, to eliminate the effects from the size of the data set and normalize the prediction performance of the GP model, the mean of the absolute value for the prediction and measurement error is considered by applying $k^{-1} \boldsymbol{1}^{\rm T}$. Therefore, $\varepsilon_{i}(t_{k})$ is a reasonable metric to evaluate the performance of the GP models without heavy variance computation.
\end{remark}

However, one must consider that the historical estimation errors can exhibit significant disparities, ranging from scenarios where an agent possesses an ideal prior knowledge resulting in zero error, to instances where an agent's prior information is grossly inaccurate, leading to exceedingly substantial errors. Consequently, there arises a necessity to standardize these errors to a suitable range. In this context, it becomes imperative to normalize them within the interval $[0, 1]$, i.e. the min-max normalization for accumulated historical estimation error, which is expressed in
\begin{align}\label{eq_tildeVarepsilon}
    \tilde{\varepsilon}_{i}(t_{k}) =  \frac{{\varepsilon}_{i}(t_{k})-{\varepsilon}_{\min,i}(t_{k})}{{\varepsilon}_{\max, i}(t_{k})-{\varepsilon}_{\min, i}(t_{k})},
\end{align}
where $ {\varepsilon}_{\max, i}(t_{k}) = \max_{i\in \bar{\mathcal{N}}_i} {\varepsilon}_{i}(t_{k})$ and $ {\varepsilon}_{\min, i}(t_{k}) = \min_{i\in \bar{\mathcal{N}}_i} {\varepsilon}_{i}(t_{k})$. The variable $\tilde{\varepsilon}_{i}(t_{k})$ serves the dual purpose of standardizing error magnitudes and facilitating threshold-based decision making. It not only simplifies the comparison and analysis of errors across different scenarios or datasets but also streamlines the establishment of thresholds for acceptable errors, which is used for designing the elective strategy.

\subsection{Prior-Aware Elective Cooperative Learning}\label{subsec_PriorAwareLearning}

Through the incorporation of the quantifiable term $\tilde{\varepsilon}$, we introduce an elective learning function denoted as $\alpha_{ij}$ leveraging the average accumulated historical estimation errors \eqref{eq_tildeVarepsilon}. Essentially, it informs us about the degree of trust in the GP models and the circumstances in which the agent $i$ requires calculations for prediction from its neighboring agent $j (j\in\bar{\mathcal{N}}_i)$, including itself. This inclusion is particularly relevant when agent $i$ seeks to calculate predictions at query point $\boldsymbol{x}_i(t_k)$ utilizing its local GP model. The elective function for agent $i$ is designed as
\begin{align}\label{eq_activeFunction}
\alpha_{ij} \left (t_{k},\bar{S}_i \right ) =  \begin{cases}
a_{ij} & \tilde{\varepsilon}_{j}(t_{k}) < \bar{\varepsilon}_{i}(t_{k})  \\
 0 & \text{ otherwise }
\end{cases}, ~ j\in \bar{\mathcal{N}}_i,
\end{align}
where $\bar{\varepsilon}_i(t_{k})$ is the $\bar{S}_i$-th largest value of the set $\{ \tilde{\varepsilon}_{i}(t_{k}) \}_{i\in \bar{\mathcal{N}}_i}$ associated with the agent $i$. 
This elective function signifies that the agent $i$ exclusively selects cooperative predictions from a subset $\mathcal{S}_i(t_k) = \{ j | \alpha_{ij}(t_k, \bar{S}_i)>0, j \in \bar{\mathcal{N}}_i\}$ of $\bar{\mathcal{N}}_i$. Specifically, the subset $\mathcal{S}_i $ defines the number of $|\mathcal{S}_i| = |\bar{\mathcal{N}_i}|- \bar{S}_i$ trustworthy agents in the set $\bar{\mathcal{N}_i}$ for the aggregation prediction. Therefore, this elective function affords the agent the capability to determine the number of collaborators, including itself, that it wishes to engage in computing joint inferences for the unknown function. 

By employing the proposed elective function in conjunction with the normalized accumulated historical error \eqref{eq_tildeVarepsilon}, we formulate the elective prior-aware aggregation weight function for the $i$-th agent designed by
\begin{align} \label{eq_omega_varepsilon}
     \tilde{\omega}^{\varepsilon}_{ij}(t_{k})= \varphi_{ij}\big( \cup_{j \in \bar{\mathcal{N}}_i} \alpha_{ij}(t_k, \bar{S}_{i}) h^{\varepsilon}(\tilde{\varepsilon}_j(t_{k}) ) \big) , ~ j\in \bar{\mathcal{N}}_i,
\end{align}
which can be simplified as 
\begin{align} \label{eq_omega_varepsilon2}
     \tilde{\omega}^{\varepsilon}_{ij}(t_{k})=\varphi_{ij}\big( \cup_{j \in {\mathcal{S}}_i}  h^{\varepsilon}(\tilde{\varepsilon}_j(t_{k}) ) \big),~ j\in {\mathcal{S}}_i,
\end{align}
according to the definition of $\mathcal{S}_i$, where $\varphi_{ij}(\cdot)$ is a proportional function
\begin{align}\label{eq_proportional}
    \varphi_{ij}(\cup_{s \in \mathcal{S}_i} w_{is}) = \frac{w_{ij}}{\sum_{s \in \mathcal{S}_i} w_{is}}.
\end{align}
Since smaller estimation error $\tilde{\varepsilon}_j(t_{k})$ indicates more reliable performance for GP model $j$ with larger $\tilde{\omega}^{\varepsilon}_{ij}(t_{k})$, the positive function $h^{\varepsilon}(\cdot): \mathbb{R}_+ \to \mathbb{R}_+$ is designed as monotonically decreasing, i.e., $\forall \tilde{\varepsilon}_i, \tilde{\varepsilon}_j \in \mathbb{R}_+$ with $\forall \tilde{\varepsilon}_i \leq \forall \tilde{\varepsilon}_j $ it holds $h^{\varepsilon}(\forall \tilde{\varepsilon}_i) \geq h^{\varepsilon}(\forall \tilde{\varepsilon}_j)$.
Moreover, when $h^{\varepsilon}(\cdot)$ is well-defined, i.e., not tend to be infinite, when the input is close to $0$.
With the above requirements, the function $h^{\varepsilon}(\cdot)$ can be designed as 
\begin{align}\label{eq_hVarepsilon}
    h^{\varepsilon}(\tilde{\varepsilon}_j(t_{k})) = \frac{\sigma_{h_i} \sqrt{2\pi}} {\exp \Big(- \frac{1}{2}\Big(\frac{\tilde{\varepsilon}_j(t_{k})-\bar{\varepsilon}_i(t_{k})} {\sigma_{h_i}}\Big)^2\Big)},
\end{align}
where the scaling factor $\sigma_{h_i}\in \mathbb{R}_+$ is the standard deviation value of the Gaussian distribution in the denominator of \cref{eq_hVarepsilon}. The rationale behind utilizing the function \eqref{eq_hVarepsilon} instead of \cref{eq_tildeVarepsilon} as the weighting scheme lies in the fact that the parameter $\sigma_{h_i}$ is a trainable variable, affording the flexibility to optimize the distribution of aggregation weights. More importantly, an additional crucial consideration is the necessity to prevent singular values from arising. Notably, the elective aggregation weight function  \cref{eq_omega_varepsilon} presents an advantageous feature wherein the computation of aggregation weights does not impose a significant computational burden, as these weights are determined based on the prior estimation errors of collaborators. However, it is well-established that the posterior variance in GPR serves as an indicator of prediction uncertainties \cite{deisenrothDistributedGaussianProcesses2015}. This metric quantifies the confidence degree of predictions with respect to the training dataset, as employed in the concept presented in \cite{yangDistributedLearningConsensus2021}. In light of this, we incorporate this notion with \cref{eq_proportional} to design the elective weight based on the variance $\tilde{\omega}^{\sigma}_{ij}(t_{k})$ of GP as \looseness=-1
\begin{align} \label{eq_omega_sigma}
    \tilde{\omega}^{\sigma}_{ij}(t_{k}) = \varphi_{ij} \big( \cup_{j \in \bar{\mathcal{N}}_i} \alpha_{ij}(t_k, \bar{S}_{i}) h^{\sigma}(\sigma_j(\boldsymbol{x}(t_k)) ) \big)
\end{align}
where $h^{\sigma}(\bullet): \mathbb{R}\to \mathbb{R}$ is $\bullet^{-2}$. Therefore, considering both the elective weights \eqref{eq_omega_varepsilon} and \eqref{eq_omega_sigma}, we combine them by using the following method
\begin{align}\label{eq_omega}
    {\omega}_{ij}(t_{k}) = \rho(\tilde{\omega}^{\varepsilon}_{ij}(t_{k}), \tilde{\omega}^{\sigma}_{ij}(t_{k})),
\end{align}
where $\rho(\cdot,\cdot): \mathbb{R} \times \mathbb{R} \to \mathbb{R}$ is designed as a function that can balance the impact between the aggregation weights based on prior estimation error and the weights based on posteriors. 
\begin{remark}
The design of the function $\rho(\cdot)$ is restraint under the condition $\sum_{j \in \bar{\mathcal{N}}_i} \tilde{\omega}_{ij}(t_{k}) = 1$. A valid choice of $\rho$ can be
\begin{align} \label{eq_rho_remark}
    \rho(\tilde{\omega}^{e}_{ij}(t_{k}), \tilde{\omega}^{\sigma}_{ij}(t_{k})) = \varphi_{ij} \Big( \cup_{j \in \bar{\mathcal{N}}_i} \big(\tilde{\omega}^{\varepsilon}_{ij}(t_{k})\big)^c \big(\tilde{\omega}^{\sigma}_{ij}(t_{k})\big)^{1-c} \Big),
\end{align}
where $0\leq c\leq 1 \in\mathbb{R}_{0,+}$ serves as a means to modulate the influence of the first and second input variables in a proportional manner. Consequently, the manipulation of the factor $c$ affords us the ability to finely adjust the relative significance of two key metrics. 
This choice of $\rho$ guarantees $\sum_{j \in \bar{\mathcal{N}}_i} {\omega}_{ij}(t_{k}) = 1$ considering the definition of function $\varphi_{ij}$ in \eqref{eq_proportional}.
\end{remark}
It is essential to acknowledge that the aggregation weights in \cref{eq_omega_sigma} entail increased computational demands on the collaborating agents, along with a higher volume of information exchange to convey the posterior variances. Nevertheless, these adjustments yield a richer source of predictive information from the collaborators. This augmentation has the potential to enhance predictions with \cref{eq_rho_remark} under fine-tuned hyperparameters. However, Pri-GP provides a valuable avenue for achieving such adaptability, particularly in situations where computational resources are constrained, considering the calculation of variance infeasible due to its inherent complexity $\mathcal{O}(N^2)$ or resulting in substantial processing delays, circumstances under which the POE method may prove ineffective. 
\begin{remark}
    Owing to the inherent characteristics of Bayesian learning methodologies, the posterior distribution continually refines itself with the assimilation of additional training data, thereby mitigating the influence of the prior distribution. Nevertheless, our approach offers a broader perspective on quantifying the model's confidence, transcending the limitations of localized query points within the training data domain. Moreover, even in scenarios where all predictive regions have been fully explored and observed, it becomes feasible to set the factor $c=0$ in \cref{eq_rho_remark}. The Pri-GP transitions into an elective POE, where the determination of aggregation weights relies solely on posterior variance. When $c=1$, it signifies that the weighting scheme exclusively relies on the prior-aware aggregation weights in \cref{eq_omega_varepsilon}. Furthermore, it is pertinent to underscore that expeditious acquisition of aggregation weights can be facilitated by bypassing the computation of variance altogether. 
\end{remark}

As the aggregation weight function defined above, \cref{eq_tilde_f} can be written as
\begin{align}\label{eq_prediction}
    \tilde{f}_i(\boldsymbol{x}(t_k)) &=  \sum_{j\in \bar{\mathcal{N}}_i}\omega_{ij}(t_{k}) \phi_{ij}\big(\boldsymbol{x}(t_k), \mathbb{D}_j, \hat{f}_j(\boldsymbol{x}(t_k))\big) \nonumber\\
    & = \sum_{j\in \bar{\mathcal{N}}_i}\omega_{ij}(t_{k}) \mu_j(\boldsymbol{x}(t_k)), ~ i \in \mathcal{V}.
\end{align}
Therefore, to obtain the aggregated prediction, the exchanged information necessitates the sharing of two critical components: firstly, the posterior mean for prediction; secondly, the cumulative historical estimation error for the computation of the aggregation weights. To facilitate a better understanding of the algorithm's operation, we provide a pseudo-code representation of Pri-GP in \cref{alg_PriGP}.
\begin{algorithm}[t]
\caption{Pri-GP Algorithm}\label{alg_PriGP}
\begin{algorithmic}
\Require $S \geq 2$  \Comment{number of agents}

\For{$i =  1 :S$}

\For{$j \in \bar{\mathcal{N}}_i$}

\State $\varepsilon_{ij}(t_{k})$, $\tilde{\varepsilon}_{ij}(t_{k})$, $\alpha_{ij}(t_k, \bar{S}_i)$ $\leftarrow$ \cref{eq_varepsilon}, \cref{eq_tildeVarepsilon}, \cref{eq_activeFunction}

\If {$\alpha_{ij}(t_k, \bar{S}_i) \ne 0$}

\If{$c=1$}
\State $\tilde{\omega}_{ij}^{\varepsilon}(t_{k})$  $\leftarrow$  \cref{eq_omega_varepsilon2}
\State Calculate ${\omega}_{ij}(t_{k}) = {\varphi}_{ij} ( \tilde{\omega}_{ij}^{\varepsilon}(t_{k}), 1 )$
\ElsIf{$c\ne1$}
\State $\tilde{\omega}_{ij}^{\sigma}(t_{k})$, ${\omega}_{ij}(t_{k}) $   $\leftarrow$  \cref{eq_omega_sigma}, \cref{eq_omega}
\EndIf
\State Calculate $\phi_{ij}\big(\boldsymbol{x}(t_k), \mathbb{D}_j, \hat{f}_j(\boldsymbol{x}(t_k))\big)$

\EndIf

\EndFor

\State $\tilde{f}_i(\boldsymbol{x}(t_k))$ $\leftarrow$\cref{eq_prediction}

\If {$y(t_k) \ne \emptyset$}
\State $e_i(\boldsymbol{x}(t_{k}))$ $\leftarrow$  \cref{eq_prior_estimation_error}
\EndIf

\EndFor
\end{algorithmic}
\end{algorithm}

\subsection{Prediction with Probabilistic Guarantee}\label{subsec_predictionErrorBound}

Before analyzing the prediction performance for the MAS with the proposed prior-aware elective distributed learning, we first quantify the prediction error bound for a single GP model with prior information, which is shown in the following lemma.
\begin{lemma} \label{lemma_single_GP_error_bound}
For an unknown function satisfying \cref{ass_unknown_f}, a GP model is given with a training data $\mathbb{D} = \{ \boldsymbol{X}, \boldsymbol{Y} \}$ set containing $N = | \mathbb{D} | \in \mathbb{N}$ samples under \cref{ass_dataset}. Moreover, choose the kernel function $\kappa(\cdot)$ satisfying \cref{ass_kernel} and a Lipschitz continuous prior mean function $\hat{f}(\cdot)$ with Lipschitz constant $L_{\hat{f}} \in \mathbb{R}_+$.
Pick the grid factor $\tau \in \mathbb{R}_+$ and $\delta \in (0,1)$,
then the prediction error with prior information is uniformly bounded by
    \begin{align} \label{eq_error_bound}
        | \mu(\boldsymbol{x}) - f(\boldsymbol{x})| \le \eta(\boldsymbol{x}) = \sqrt{\beta} \sigma(\boldsymbol{x}) + \gamma \tau, \forall \boldsymbol{x} \in \mathbb{X}
    \end{align}
    with a probability of at least $1 - \delta$, where $\gamma = L_f + L_{\hat{f}} +  \sqrt{\beta L_{\sigma^2}\tau} + \sqrt{N} L_{k}  \| \boldsymbol{K}(\boldsymbol{X})^{-1} \big(\boldsymbol{Y}_k^{\rm T} - \hat{\boldsymbol{f}}(\boldsymbol{X})^{\rm T}\big) \|$ and the Lipschitz constant of the posterior variance $L_{\sigma^2} = 2 L_{\kappa}\big(1+N\big\|\boldsymbol{K}(\boldsymbol{X})^{-1}\big\| \max _{x, x^{\prime} \in \mathbb{X}} k\left(x, x^{\prime}\right)\big)$.
    The constant $\beta=  2 \sum_{j=1}^{m} \log{\big( \frac{\sqrt{m}} {2 \tau} (\bar{x}_{j} - \underline{x}_{j}) + 1 \big)} - 2\log{\delta}$, where $\bar{x}_{j}$ and $\underline{x}_{j}$ denote the maximum and minimum of the $j$-th dimension of $\boldsymbol{x}$ in the domain $\mathbb{X}$, i.e., $\bar{x}_{j} = \max_{\boldsymbol{x} \in \mathbb{X}} x_j$ and $\underline{x}_{j} = \min_{\boldsymbol{x} \in \mathbb{X}} x_j$.

\end{lemma}
\begin{proof}
To prove the uniform error bound in $\mathbb{X}$, we first define the discrete domain $\mathbb{X}_\tau$ based on the grid factor $\tau$, such that for each element $\boldsymbol{x} \in \mathbb{X}$ there exists an element $\boldsymbol{x}' \in \mathbb{X}_\tau$ satisfying $\| \boldsymbol{x} - \boldsymbol{x}' \| < \tau$.
The domain $\mathbb{X}_\tau$ is finite, whose cardinality is bounded according to \cite{dai2023can} as $| \mathbb{X}_\tau | \le \prod_{j = 1}^m \left( \frac{\sqrt{m}} {2 \tau} (\bar{x}_{j} - \underline{x}_{j}) + 1 \right)$.
Moreover, employing Lemma 5.1 in \cite{srinivasInformationTheoreticRegretBounds2012a}, the uniform error bound within $\mathbb{X}_\tau$ is written as
\begin{align}
    \Pr\{ | \mu(\boldsymbol{x}') - f(\boldsymbol{x}') | \le \sqrt{\beta} \sigma(\boldsymbol{x}'), \forall \boldsymbol{x}' \in \mathbb{X}_\tau \} \ge 1 - \delta
\end{align}
considering $\beta = 2 \log(| \mathbb{X}_\tau | / \delta)$.
Then, due to fact that $\| \boldsymbol{x} - \boldsymbol{x}'\| \le \tau$, the prediction error within the domain $\mathbb{X}$ is bounded by
    \begin{align} \label{eq_eta_half}
        &| \mu(\boldsymbol{x}) - f(\boldsymbol{x}) | \le | \mu(\boldsymbol{x}) - \mu(\boldsymbol{x}') | + | f(\boldsymbol{x}) - f(\boldsymbol{x}') | + | \mu(\boldsymbol{x}') - f(\boldsymbol{x}') | \nonumber \\
        & \le \sqrt{\beta} \sigma(\boldsymbol{x}') + L_f \tau + | \mu(\boldsymbol{x}) - \mu(\boldsymbol{x}') | \nonumber\\
        &\le \sqrt{\beta} \sigma(\boldsymbol{x}) + \sqrt{\beta L_{\sigma}\tau}+ L_f\tau  + | \mu(\boldsymbol{x}) - \mu(\boldsymbol{x}') | 
    \end{align}
    for all $\boldsymbol{x} \in \mathbb{X}$.
While the Lipschitz constant for the posterior mean additionally depends on the prior mean, i.e.,
    \begin{align} \label{eq_L_mu}
        &| \mu(\boldsymbol{x}) - \mu(\boldsymbol{x}') | \le | \big( K (\boldsymbol{x}, \boldsymbol{X} ) -  K (\boldsymbol{x}', \boldsymbol{X} ) \big) \boldsymbol{K}(\boldsymbol{X})^{-1} \big(\boldsymbol{Y}_k^{\rm T} - \hat{\boldsymbol{f}}(\boldsymbol{X})^{\rm T}\big) | \nonumber\\
        & \qquad\qquad\qquad \quad +|\hat{f}(\boldsymbol{x}) - \hat{f}(\boldsymbol{x}')|   \\
        &\le \sqrt{N} L_{k}  \| \boldsymbol{K}(\boldsymbol{X})^{-1} \big(\boldsymbol{Y}^{\rm T} - \hat{\boldsymbol{f}}(\boldsymbol{X})^{\rm T}\big) \| \| \boldsymbol{x} - \boldsymbol{x}'\| +L_{\hat{f}} \| \boldsymbol{x} - \boldsymbol{x}' \|. \nonumber
    \end{align}
    Apply \eqref{eq_L_mu} into \eqref{eq_eta_half}, then the uniform error bound in \eqref{eq_error_bound} for $\mathbb{X}$ is derived, which completes the proof.
\end{proof}

Based on \cref{lemma_single_GP_error_bound}, we derive the overall prediction error bound of the MAS.
\begin{theorem}
Consider a MAS with $S$ agents using the Pri-GP algorithm to infer the unknown function $f(\cdot)$ under \cref{ass_unknown_f}.
Equip a GP model on each agent $i$ with the kernel function satisfying \cref{ass_kernel} and a Lipschitz continuous prior $\hat{f}(\cdot)$ with Lipschitz constant $L_{\hat{f},i}$.
Moreover, let each agent has its individual data set $\mathbb{D}_i$ satisfying \cref{ass_dataset}. Pick $\tau \in \mathbb{R}_+$ and $\delta \in (0,1)$
, then the overall prediction error denotes
    \begin{align} \label{eq_overall_prediction_error_bound}
        \| \tilde{\boldsymbol{f}}(\boldsymbol{x}) - \boldsymbol{f}(\boldsymbol{x}) \| \le \| [\tilde{\eta}_1(\boldsymbol{x}_1), \cdots, \tilde{\eta}_S(\boldsymbol{x}_S)]^T\|,
    \end{align}
    with probability of at least $1 - \sum_{i=1}^S |{\mathcal{S}}_i| \delta$, where the aggregated function and prediction denote $\tilde{\boldsymbol{f}}(\boldsymbol{x}) = [\tilde{f}_1(\boldsymbol{x}_1), \cdots, \tilde{f}_S(\boldsymbol{x}_S)]^T$, $\boldsymbol{f}(\boldsymbol{x}) = [f(\boldsymbol{x}_1), \cdots, f(\boldsymbol{x}_S)]^T$, and
    \begin{align} \label{eq_single_aggregated_error_bound}
        \tilde{\eta}_i(\boldsymbol{x}_i) = \frac{\sum_{j \in {\mathcal{S}}_i}  h_{\varepsilon}(\tilde{\varepsilon}_j(t_{k}) )^c h_{\sigma}(\sigma_j(\boldsymbol{x}(t_k))^{1-c} \eta_j(\boldsymbol{x}_i) }{\sum_{j \in {\mathcal{S}}_i}h_{\varepsilon}(\tilde{\varepsilon}_j(t_{k}) )^c h_{\sigma}(\sigma_j(\boldsymbol{x}(t_k))^{1-c}}.
    \end{align}
\end{theorem}
\begin{proof}
    Due to the property of function $\varphi(\cdot)$, the prediction error for the $i$-th agent is written as
    \begin{align} \label{eq_ftilde_f}
        &| \tilde{f}_i(\boldsymbol{x}_i(t_{k})) - f(\boldsymbol{x}_i(t_{k})) | = \Big|\sum_{j \in \bar{\mathcal{N}}_i} {\omega}_{ij}(t_{k}) \big(\mu_i(\boldsymbol{x}(t_k)) - f(\boldsymbol{x}_i(t_{k})) \big) \Big| \nonumber \\
        &\le \sum_{j \in \bar{\mathcal{N}}_i} {\omega}_{ij}(t_{k}) |\mu_j(\boldsymbol{x}(t_k)) - f(\boldsymbol{x}_i(t_{k}))| \le \sum_{j \in \bar{\mathcal{N}}_i} {\omega}_{ij}(t_{k}) \eta_j(\boldsymbol{x}_i(t_k)),
    \end{align}
    where the second inequality is derived from \cref{lemma_single_GP_error_bound}.
    Consider the definition of ${\omega}_{ij}(t_{k})$ in \eqref{eq_omega}, the aggregation weight is rewritten as
    \begin{align}
        {\omega}_{ij}(t_{k}) = \frac{(\tilde{\omega}^{e}_{ij}(t_{k}))^c (\tilde{\omega}^{\sigma}_{ij}(t_{k}))^{1-c}}{\sum_{s \in \bar{\mathcal{N}}_i} (\tilde{\omega}^{e}_{is}(t_{k}))^c (\tilde{\omega}^{\sigma}_{is}(t_{k}))^{1-c}}
    \end{align}
With \eqref{eq_omega_varepsilon} and \eqref{eq_omega_sigma}, one has
    \begin{align} \label{eq_we_c_wsigma_1c}
        &(\tilde{\omega}^{e}_{ij}(t_{k}))^c (\tilde{\omega}^{\sigma}_{ij}(t_{k}))^{1-c} \\
        &\qquad \qquad= \frac{\alpha_{ij}(t_k, \bar{S}_i) h_{\varepsilon}(\tilde{\varepsilon}_j(t_{k}))^c h_{\sigma}(\sigma_j(\boldsymbol{x}(t_k)) )^{1-c}}{\big(\sum_{s \in {\mathcal{S}}_i} h_{\varepsilon}(\tilde{\varepsilon}_j(t_{k}) )\big)^c \big(\sum_{s \in {\mathcal{S}}_i}  h_{\sigma}(\sigma_j(\boldsymbol{x}(t_k)) )\big)^{1-c}}. \nonumber
    \end{align}
    Apply \eqref{eq_we_c_wsigma_1c} into \eqref{eq_ftilde_f}, then the result in \eqref{eq_single_aggregated_error_bound} is derived with the probability of at least $1 - |{\mathcal{S}}_i| \delta$ using union bound.
    Moreover, employing union bound again for different agents, the overall prediction error bound in \eqref{eq_overall_prediction_error_bound} is obtained.
\end{proof}

\section{Numerical Evaluation}
To effectively elucidate the efficacy of our proposed algorithms, we commence by employing an approximated sine function as a demonstrative vehicle expounding upon the fundamental principles of Pri-GP, as explicated in \cref{subsec_functionApproximation}. Furthermore, we showcase the proficiency of our novel algorithms in identifying the dynamics characterizing autonomous systems in \cref{subsec_systemIdentification}.

\subsection{Function Approximation}\label{subsec_functionApproximation}
In this subsection, we investigate the MAS comprising 4 agents, each equipped with an identical dataset but possessing distinct prior knowledge represented as $\hat{f}$. The rationale behind this experiment is to facilitate an in-depth analysis of the impact of varying prior knowledge on predictions when employing an individual learning strategy. Furthermore, we aim to draw comparisons with different distributed learning methodologies showing the proposed algorithms are superior to others. 

\begin{figure}[t]
  \centering
  \includegraphics[width=1\linewidth]{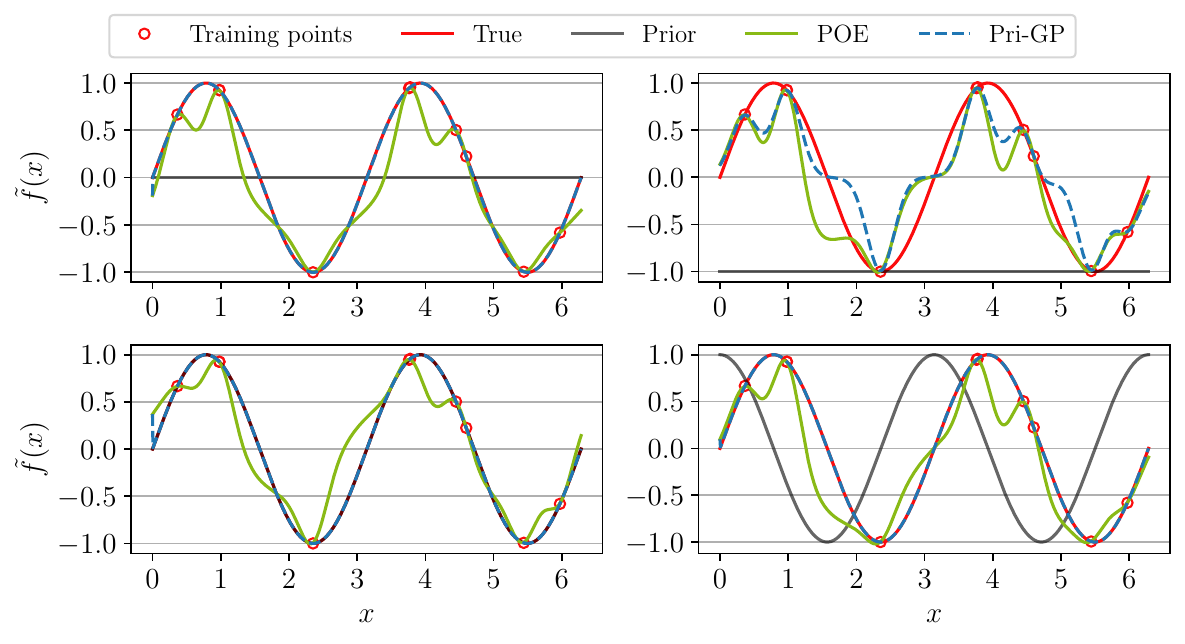}
  \caption{True, prior and posterior curves.}
  \label{fig_f4}
\end{figure}

The target function for approximation in this scenario is chosen as $sin(2x)$. 
We set the prior function of agent $1$ to $\hat{f}_1(x) = 0$ considering the agent does not have any knowledge of the unknown function, which is a general assumption. Moreover, let agent $3$ have the accurate function $\hat{f}_3(x) = sin(2x)$ as the target function and the $2$-nd agent and the $4$-th agent as $\hat{f}_2(x) = -1$ and $\hat{f}_4(x) = cos(2x)$, respectively. The adjacency matrix of the communication graph of this MAS is
\begin{align*}
  \mathcal{A} =   \begin{bmatrix}
 1 & 1 & 1 & 0\\
 1 &  1& 0 &1 \\
 1 & 0 & 1 & 1\\
 0 & 1 & 1 & 1
\end{bmatrix}.
\end{align*}
Let $\bar{S}_1 =  \bar{S}_2 = \bar{S}_3 = \bar{S}_4 =2$, and the $8$ identical training input are randomly selected obeying uniform distribution over the range $[ 0, 2\pi )$. And set the hyperparameters of the kernel function are chosen as $\sigma_{r}=1, l_{j}=0.2, j= 1,2,3$ and the noise variance of the noise is $\sigma_n = 0.1$ for all agents. The curves presented in \cref{fig_f4} show the results of function approximation. It becomes apparent that the curve using individual learning with Gaussian Processes (IGP), i.e., the agent predicts the unknown function independently, closely approximates the prior function in the absence of training data. This observation underscores the significant influence of prior knowledge on predictions when no data are available. 

\begin{figure}[h]
  \centering
  \includegraphics[width=1\linewidth]{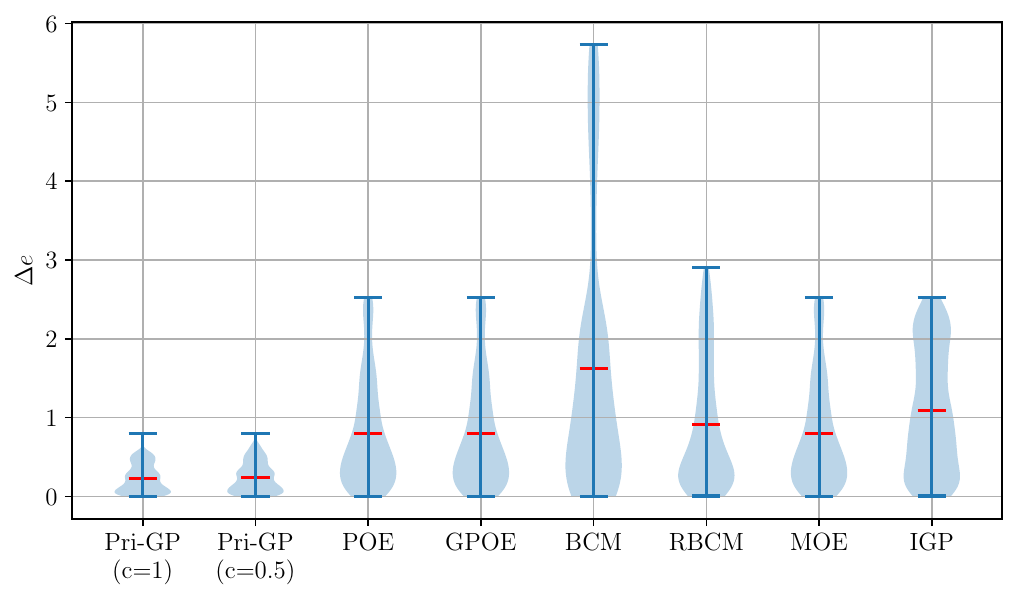}
  \caption{Violin plots of prediction errors for different methods. The red line is the mean value and the top/bottom horizontal blue bar is the maximal/minimal value.}
  \label{fig_errorsFunctionApproximation}
\end{figure}
\begin{figure*}[t]
    \begin{subfigure}{0.31\linewidth} 
        \centering
        \includegraphics[width=1\linewidth]{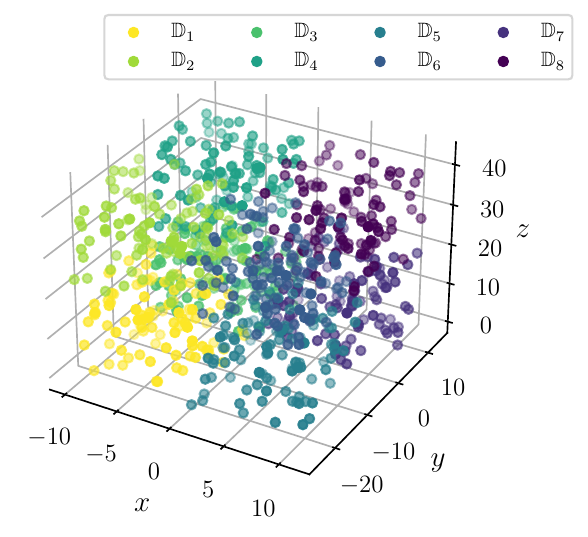}
        \caption{Training data}
        \label{fig_8dataset}
    \end{subfigure}%
    \begin{subfigure}{0.685\linewidth} 
        \centering
        \includegraphics[width=1\linewidth]{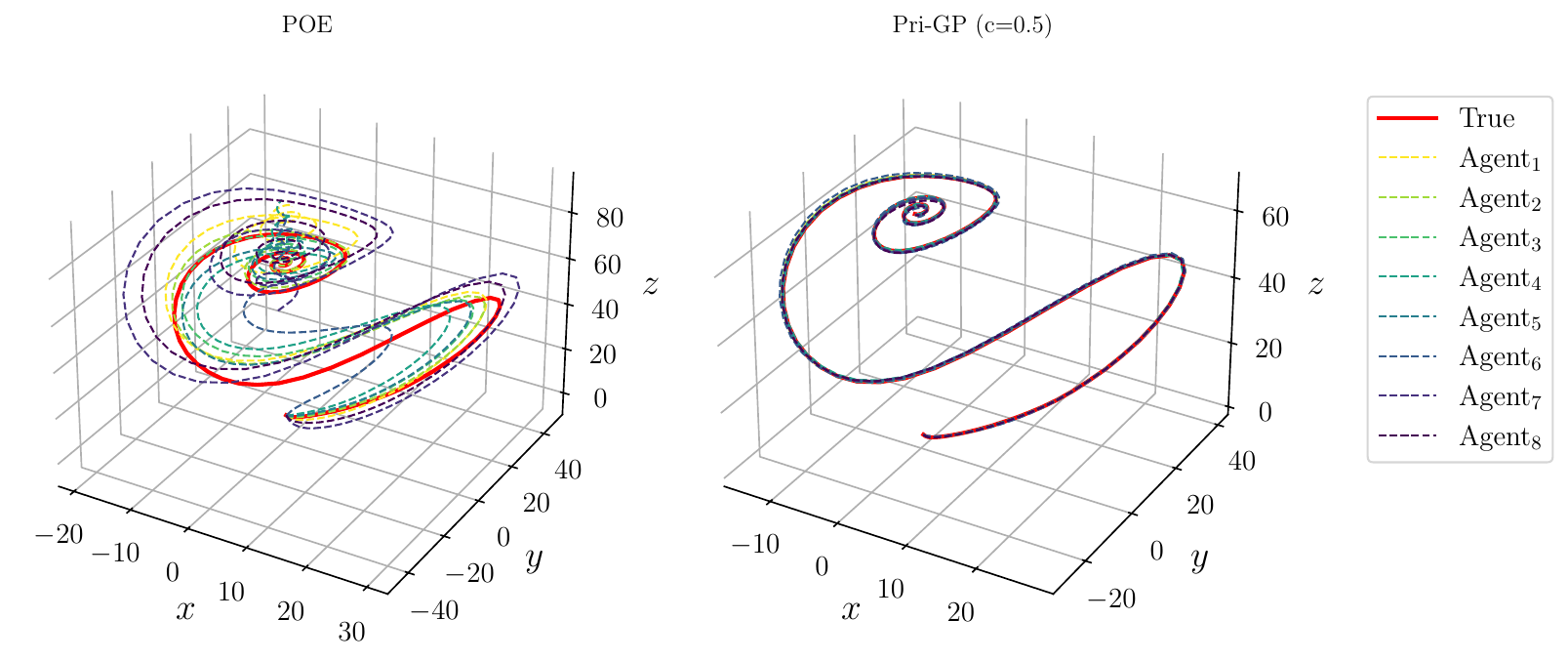}
        \caption{System trajectories}
        \label{fig_2GPs}
    \end{subfigure}%
\caption{3 Dimension Plots}
    \label{fig_dyn}
\end{figure*}
To facilitate a more nuanced comparison of our proposed methods with existing approaches, we provide the prediction error, denoted by $\Delta e = |\tilde{f}(x) - f(x)|$, of all agents with violin plots in \cref{fig_errorsFunctionApproximation} to analyze the distribution of prediction errors. Furthermore, the average of the $1000$ prediction errors regarding each agent is illustrated \cref{tab_functionApproximation}. It is evident that Pri-GP methods outperform the other methods, which have the lowest sum of average prediction errors. Notably, while the overall prediction errors for the MAS letting $c\ne 1$ may appear less favorable when compared to $c=1$, a closer examination reveals that agent $2$, in particular, benefits from Pri-GP with $c=0.5$. 
The BCM method manifests heightened errors that can be attributed to that BCM tends to disproportionately accentuate the influence of prior variance within the aggregation weights. Moreover, it is crucial to note that the similarity in results of the POE, POE, and GPOE methods arises from the identical training datasets employed by all agents. In order to explore the impact of distinct training datasets, a more intricate scenario is examined in \cref{subsec_systemIdentification}. \looseness=-1

\begin{table}[t]
	\caption{Average prediction errors ($\times 10^{-2}$)}
	\label{tab_functionApproximation}
	\begin{tabular}{lccccc}\toprule
		\textit{Methods} & \textit{Agent 1} & \textit{Agent 2}& \textit{Agent 3}& \textit{Agent 4}& \textit{Sum} \\ \midrule
		Pri-GP ($c\!=\!1$)      & \textbf{0.044} & 22.25 & {0.048} & \textbf{0.032}& \textbf{22.37}\\
		Pri-GP ($c\!=\!\frac{1}{2}$) & 0.734 & \textbf{22.13}& 0.515 & 0.958& 24.34\\
        POE         & 16.64 & 24.24& 17.83 & 20.87& 79.58\\
        GPOE         & 16.64 & 24.24& 17.83 & 20.87& 79.58\\
        BCM         & 38.48 & 41.12& 37.38 & 45.09& 162.07\\
        RBCM         & 22.26 & 25.31& 21.08 & 23.11& 91.76\\
        MOE         & 16.64 & 24.24& 17.82 & 20.87& 79.58\\
		IGP         & 22.26 & 42.70& \textbf{0.002} & 43.65& 108.61\\\bottomrule
	\end{tabular}
\end{table}

\subsection{Dynamical System Identification}\label{subsec_systemIdentification}
\begin{figure}[ht]
    \centering
    \includegraphics[width=0.8\linewidth]{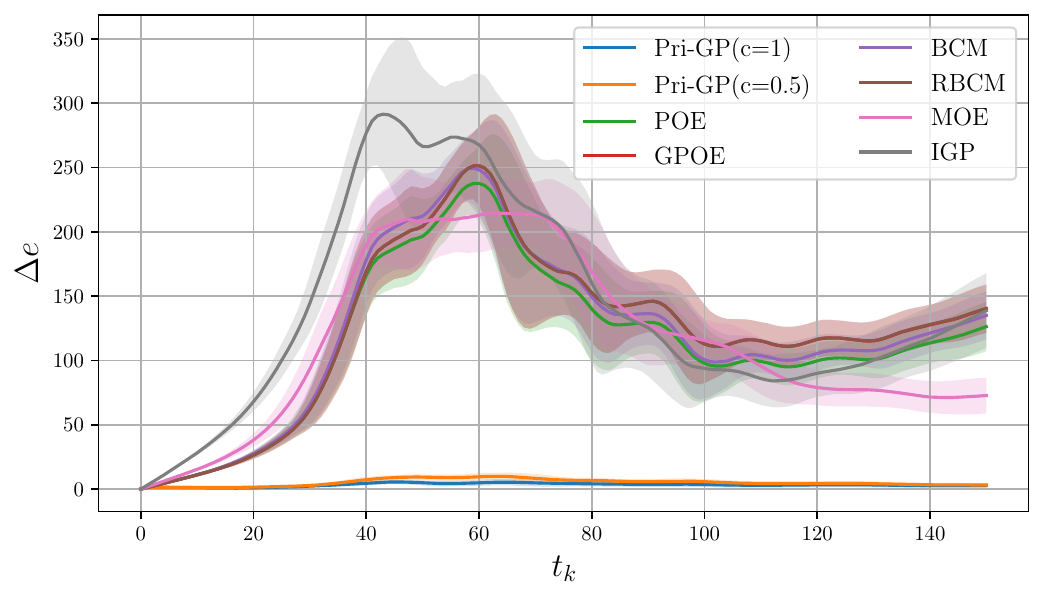}
    \caption{Mean prediction error with standard deviation}
    \label{fig_errorDynamical}
\end{figure}

To further demonstrate the capability of Pri-GP in the identification of dynamical systems, this subsection endeavors to exemplify the performance of Pri-GP with a $3$ dimensional nonlinear system
\begin{align*}
    \dot{x} &= s(y - x), \\
    \dot{y} &= rx - y - xz, \\
    \dot{z} &= xy \underbrace{-10\sin(z) - 10x - 0.5(1+\exp(-xy/10))^{-1}}_{f(\chi)},
\end{align*}
where we assume the unknown component of the system as represented by $f(\chi): \mathbb{R}^3 \to \mathbb{R}$, where $\chi = [x, y, z]$\footnote{For complete results of the simulations, refer to the extended version~\cite{yangAAMAS2024}.}. The MAS comprises $8$ agents, each equipped with unique datasets with $100$ training data satisfying the conditions specified in  \cref{ass_dataset} randomly distributed in the space $[-10, 20]\times [-25, 30] \times [0, 60]$  (see \cref{fig_8dataset}). All agents begin in the same initial states, which are randomly determined within the range $[0, 1]$. These simulations are conducted 100 times for Monte-Carlo simulations, with a time step of $0.01$, and each simulation continues for $150$ time steps. The hyperparameters of the kernel function are chosen as $\sigma_{r}=1, l_{j}=1000, j= 1,2,3$ and the noise variance of the noise is $\sigma_n = 0.1$. Furthermore, the selection of diverse prior functions in the MAS are as follows
\begin{align*}
    \hat{f}_1(\chi) &=  \hat{f}_3(\chi) =- 10sin(z) - 10x - 0.5(1+\exp(-xy/10))^{-1}, \\
    \hat{f}_2(\chi) &= 0, \qquad \hat{f}_4(\chi) = -10sin(z),  \qquad \hat{f}_5(\chi) = - 10x,\\
    \hat{f}_6(\chi) &= 10y - 0.5(1+\exp(-xy/10))^{-1},
\end{align*}
\begin{align*}
    \hat{f}_7(\chi) &= - 0.5(1+\exp(-xy/10))^{-1},\\
    \hat{f}_8(\chi) &= -10cos(z).
\end{align*}
Similar to \cref{subsec_functionApproximation}, each agent discharges one neighbor leading $\bar{S}_1 = \bar{S}_3 =4 $, $\bar{S}_2 =  \bar{S}_4 = \bar{S}_5 = \bar{S}_8 =3$, and $\bar{S}_6 = \bar{S}_7 = 2$. 
The trajectories of the dynamical system using POE and Pri-GP $(c=0.5)$ for one trail are depicted showcasing the spatial states of agents in \cref{fig_2GPs}, where the initial state is $[0, 1, 1.05]$. In the case of POE, there is no agent that closely follows the true system trajectory, while Pri-GP $(c=0.5)$ enables all agents to accurately identify the system dynamics. Only the trajectory of agent 8 slightly differs from the trajectory of the real system. This difference in performance can be attributed to Pri-GP having an inferior prior function and different neighbors. Additionally, it may be influenced by the fact that the training data points of agent $8$ are barely aligned with the true trajectory. 
\cref{fig_errorDynamical} presents a comparative analysis demonstrating the superior performance of the two Pri-GP methods across the entire experimental process. The solid lines represent the mean predictions obtained from 100 simulations, while the light-shaded areas denote the standard deviation for each method. It is evident that all alternative approaches exhibit comparable large prediction errors. 

\section{Conclusion}
In summary, Pri-GP emerges as a robust and promising solution for enhancing distributed cooperative learning within MASs. It introduces a novel approach that not only significantly improves prediction accuracy but also addresses the computational burden is distributed GPR by empowering agents to selectively request predictions from trusted neighbors. It offers several advantages, including improved prediction accuracy, reduced computational complexity, and the establishment of prediction error bounds, making it a valuable tool for applications where trustworthiness and reliability are paramount. The simulation results support the efficacy of Pri-GP, underscoring its superiority over existing methods in various scenarios, thus validating its potential utility for advancing the capabilities of MASs across a spectrum of domains, from safety-critical systems to resource-efficient distributed networks.



\begin{acks}
This work has been financially supported by the Germany Federal Ministry of Health (BMG) under grant No. 2523DAT400 (project ``AI-assisted analysis and visualization of pandemic situations'' | AI-DAVis-PANDEMICS), 
by the Federal Ministry of Education and Research of Germany in the programme of ``Souverän. Digital. Vernetzt.'' under joint project 6G-life with project identification number: 16KISK002, and by the European Research Council (ERC) Consolidator Grant ``Safe data-driven control for human-centric systems (CO-MAN)'' under grant agreement number 864686.
\end{acks}


\balance
\bibliographystyle{ACM-Reference-Format} 
\bibliography{ref}


\begin{thebibliography}{33}


\ifx \showCODEN    \undefined \def \showCODEN     #1{\unskip}     \fi
\ifx \showDOI      \undefined \def \showDOI       #1{#1}\fi
\ifx \showISBNx    \undefined \def \showISBNx     #1{\unskip}     \fi
\ifx \showISBNxiii \undefined \def \showISBNxiii  #1{\unskip}     \fi
\ifx \showISSN     \undefined \def \showISSN      #1{\unskip}     \fi
\ifx \showLCCN     \undefined \def \showLCCN      #1{\unskip}     \fi
\ifx \shownote     \undefined \def \shownote      #1{#1}          \fi
\ifx \showarticletitle \undefined \def \showarticletitle #1{#1}   \fi
\ifx \showURL      \undefined \def \showURL       {\relax}        \fi
\providecommand\bibfield[2]{#2}
\providecommand\bibinfo[2]{#2}
\providecommand\natexlab[1]{#1}
\providecommand\showeprint[2][]{arXiv:#2}

\bibitem[\protect\citeauthoryear{Alotaibi, Alqefari, and Koubaa}{Alotaibi
  et~al\mbox{.}}{2019}]%
        {8695011}
\bibfield{author}{\bibinfo{person}{Ebtehal~Turki Alotaibi},
  \bibinfo{person}{Shahad~Saleh Alqefari}, {and} \bibinfo{person}{Anis
  Koubaa}.} \bibinfo{year}{2019}\natexlab{}.
\newblock \showarticletitle{LSAR: Multi-UAV Collaboration for Search and Rescue
  Missions}.
\newblock \bibinfo{journal}{\emph{IEEE Access}}  \bibinfo{volume}{7}
  (\bibinfo{year}{2019}), \bibinfo{pages}{55817--55832}.
\newblock
\urldef\tempurl%
\url{https://doi.org/10.1109/ACCESS.2019.2912306}
\showDOI{\tempurl}


\bibitem[\protect\citeauthoryear{Cao and Fleet}{Cao and Fleet}{2015}]%
        {cao2015generalized}
\bibfield{author}{\bibinfo{person}{Yanshuai Cao} {and}
  \bibinfo{person}{David~J. Fleet}.} \bibinfo{year}{2015}\natexlab{}.
\newblock \bibinfo{title}{Generalized Product of Experts for Automatic and
  Principled Fusion of Gaussian Process Predictions}.
\newblock
\newblock
\showeprint[arxiv]{1410.7827}~[cs.LG]


\bibitem[\protect\citeauthoryear{Chowdhury and Gopalan}{Chowdhury and
  Gopalan}{[n.d.]}]%
        {chowdhuryKernelizedMultiarmedBandits2017}
\bibfield{author}{\bibinfo{person}{Sayak~Ray Chowdhury} {and}
  \bibinfo{person}{Aditya Gopalan}.} \bibinfo{year}{[n.d.]}\natexlab{}.
\newblock \showarticletitle{On {{Kernelized Multi-armed Bandits}}}. In
  \bibinfo{booktitle}{\emph{Proceedings of the 34th {{International
  Conference}} on {{Machine Learning}}}} (2017-07-17).
  \bibinfo{publisher}{{PMLR}}, \bibinfo{pages}{844--853}.
\newblock
\showISSN{2640-3498}
\urldef\tempurl%
\url{https://proceedings.mlr.press/v70/chowdhury17a.html}
\showURL{%
\tempurl}


\bibitem[\protect\citeauthoryear{Dai, Xie, and Chen}{Dai et~al\mbox{.}}{2019}]%
        {daiEventTriggeredDistributedCooperative2019}
\bibfield{author}{\bibinfo{person}{Hao Dai}, \bibinfo{person}{Jin Xie}, {and}
  \bibinfo{person}{Weisheng Chen}.} \bibinfo{year}{2019}\natexlab{}.
\newblock \showarticletitle{Event-Triggered Distributed Cooperative Learning
  Algorithms over Networks via Wavelet Approximation}.
\newblock \bibinfo{journal}{\emph{Neural Process. Lett.}} \bibinfo{volume}{50},
  \bibinfo{number}{1} (\bibinfo{date}{aug} \bibinfo{year}{2019}),
  \bibinfo{pages}{669–700}.
\newblock
\showISSN{1370-4621}
\urldef\tempurl%
\url{https://doi.org/10.1007/s11063-019-10031-x}
\showDOI{\tempurl}


\bibitem[\protect\citeauthoryear{Dai, He, Ma, and Yuan}{Dai
  et~al\mbox{.}}{2021}]%
        {daiDistributedCooperativeLearning2021}
\bibfield{author}{\bibinfo{person}{Shi-Lu Dai}, \bibinfo{person}{Shude He},
  \bibinfo{person}{Yufei Ma}, {and} \bibinfo{person}{Chengzhi Yuan}.}
  \bibinfo{year}{2021}\natexlab{}.
\newblock \showarticletitle{Distributed Cooperative Learning Control of
  Uncertain Multiagent Systems With Prescribed Performance and Preserved
  Connectivity}.
\newblock \bibinfo{journal}{\emph{IEEE Transactions on Neural Networks and
  Learning Systems}} \bibinfo{volume}{32}, \bibinfo{number}{7}
  (\bibinfo{year}{2021}), \bibinfo{pages}{3217--3229}.
\newblock
\urldef\tempurl%
\url{https://doi.org/10.1109/TNNLS.2020.3010690}
\showDOI{\tempurl}


\bibitem[\protect\citeauthoryear{Dai, Lederer, Yang, and Hirche}{Dai
  et~al\mbox{.}}{2023}]%
        {dai2023can}
\bibfield{author}{\bibinfo{person}{Xiaobing Dai}, \bibinfo{person}{Armin
  Lederer}, \bibinfo{person}{Zewen Yang}, {and} \bibinfo{person}{Sandra
  Hirche}.} \bibinfo{year}{2023}\natexlab{}.
\newblock \showarticletitle{Can Learning Deteriorate Control? Analyzing
  Computational Delays in Gaussian Process-Based Event-Triggered Online
  Learning}. In \bibinfo{booktitle}{\emph{Proceedings of The 5th Annual
  Learning for Dynamics and Control Conference}}
  \emph{(\bibinfo{series}{Proceedings of Machine Learning Research},
  Vol.~\bibinfo{volume}{211})}, \bibfield{editor}{\bibinfo{person}{Nikolai
  Matni}, \bibinfo{person}{Manfred Morari}, {and} \bibinfo{person}{George~J.
  Pappas}} (Eds.). \bibinfo{publisher}{PMLR}, \bibinfo{pages}{445--457}.
\newblock
\urldef\tempurl%
\url{https://proceedings.mlr.press/v211/dai23a.html}
\showURL{%
\tempurl}


\bibitem[\protect\citeauthoryear{Dai, Yang, and Hirche}{Dai
  et~al\mbox{.}}{2024a}]%
        {dai2024cooperative}
\bibfield{author}{\bibinfo{person}{Xiaobing Dai}, \bibinfo{person}{Zewen Yang},
  {and} \bibinfo{person}{Sandra Hirche}.} \bibinfo{year}{2024}\natexlab{a}.
\newblock \bibinfo{title}{Cooperative Online Learning for Multi-Agent System
  Control via Gaussian Processes with Event-Triggered Mechanism: Extended
  Version}.
\newblock
\newblock
\showeprint[arxiv]{2304.05138}~[eess.SY]
\urldef\tempurl%
\url{https://arxiv.org/abs/2304.05138v2}
\showURL{%
\tempurl}


\bibitem[\protect\citeauthoryear{Dai, Yang, Mengtian~Xu, Hattab, and
  Hirche}{Dai et~al\mbox{.}}{2024b}]%
        {dai2024DecentralizedEventTriggered}
\bibfield{author}{\bibinfo{person}{Xiaobing Dai}, \bibinfo{person}{Zewen Yang},
  \bibinfo{person}{Fangzhou~Liu Mengtian~Xu}, \bibinfo{person}{Georges Hattab},
  {and} \bibinfo{person}{Sandra Hirche}.} \bibinfo{year}{2024}\natexlab{b}.
\newblock \bibinfo{title}{Decentralized Event-Triggered Online Learning for
  Safe Consensus of Multi-Agent Systems with Gaussian Process Regression}.
\newblock
\newblock


\bibitem[\protect\citeauthoryear{Deisenroth and Ng}{Deisenroth and Ng}{2015}]%
        {deisenrothDistributedGaussianProcesses2015}
\bibfield{author}{\bibinfo{person}{Marc Deisenroth} {and}
  \bibinfo{person}{Jun~Wei Ng}.} \bibinfo{year}{2015}\natexlab{}.
\newblock \showarticletitle{Distributed Gaussian Processes}. In
  \bibinfo{booktitle}{\emph{Proceedings of the 32nd International Conference on
  Machine Learning}} \emph{(\bibinfo{series}{Proceedings of Machine Learning
  Research}, Vol.~\bibinfo{volume}{37})},
  \bibfield{editor}{\bibinfo{person}{Francis Bach} {and} \bibinfo{person}{David
  Blei}} (Eds.). \bibinfo{publisher}{PMLR}, \bibinfo{address}{Lille, France},
  \bibinfo{pages}{1481--1490}.
\newblock
\urldef\tempurl%
\url{https://proceedings.mlr.press/v37/deisenroth15.html}
\showURL{%
\tempurl}


\bibitem[\protect\citeauthoryear{Gao, Chen, Li, Li, and Xu}{Gao
  et~al\mbox{.}}{2020}]%
        {gaoNeuralNetworkBasedDistributed2020}
\bibfield{author}{\bibinfo{person}{Fei Gao}, \bibinfo{person}{Weisheng Chen},
  \bibinfo{person}{Zhiwu Li}, \bibinfo{person}{Jing Li}, {and}
  \bibinfo{person}{Bin Xu}.} \bibinfo{year}{2020}\natexlab{}.
\newblock \showarticletitle{Neural Network-Based Distributed Cooperative
  Learning Control for Multiagent Systems via Event-Triggered Communication}.
\newblock \bibinfo{journal}{\emph{IEEE Transactions on Neural Networks and
  Learning Systems}} \bibinfo{volume}{31}, \bibinfo{number}{2}
  (\bibinfo{year}{2020}), \bibinfo{pages}{407--419}.
\newblock
\urldef\tempurl%
\url{https://doi.org/10.1109/TNNLS.2019.2904253}
\showDOI{\tempurl}


\bibitem[\protect\citeauthoryear{Jafari and Xu}{Jafari and Xu}{2018}]%
        {jafariIntelligentControlUnmanned2018}
\bibfield{author}{\bibinfo{person}{Mohammad Jafari} {and} \bibinfo{person}{Hao
  Xu}.} \bibinfo{year}{2018}\natexlab{}.
\newblock \showarticletitle{Intelligent Control for Unmanned Aerial Systems
  with System Uncertainties and Disturbances Using Artificial Neural Network}.
\newblock \bibinfo{journal}{\emph{Drones}} \bibinfo{volume}{2},
  \bibinfo{number}{3} (\bibinfo{year}{2018}).
\newblock
\showISSN{2504-446X}
\urldef\tempurl%
\url{https://doi.org/10.3390/drones2030030}
\showDOI{\tempurl}


\bibitem[\protect\citeauthoryear{Khalil}{Khalil}{2002}]%
        {khalil2015nonlinear}
\bibfield{author}{\bibinfo{person}{Hassan~K Khalil}.}
  \bibinfo{year}{2002}\natexlab{}.
\newblock \bibinfo{booktitle}{\emph{{Nonlinear Systems}}}.
\newblock \bibinfo{publisher}{Prentice-Hall}.
\newblock


\bibitem[\protect\citeauthoryear{Kim, Chang, and Shim}{Kim
  et~al\mbox{.}}{[n.d.]}]%
        {kimModelReferenceGaussian2023}
\bibfield{author}{\bibinfo{person}{Hyuntae Kim}, \bibinfo{person}{Hamin Chang},
  {and} \bibinfo{person}{Hyungbo Shim}.} \bibinfo{year}{[n.d.]}\natexlab{}.
\newblock \bibinfo{booktitle}{\emph{Model {{Reference Gaussian Process
  Regression}}: {{Data-Driven State Feedback Controller}}}}.
\newblock
\urldef\tempurl%
\url{https://doi.org/10.48550/arXiv.2303.09828}
\showDOI{\tempurl}
\showeprint[arxiv]{2303.09828}~[cs, eess]


\bibitem[\protect\citeauthoryear{Lederer, Umlauft, and Hirche}{Lederer
  et~al\mbox{.}}{2019}]%
        {ledererUniformErrorBounds2019}
\bibfield{author}{\bibinfo{person}{Armin Lederer}, \bibinfo{person}{Jonas
  Umlauft}, {and} \bibinfo{person}{Sandra Hirche}.}
  \bibinfo{year}{2019}\natexlab{}.
\newblock \showarticletitle{Uniform Error Bounds for Gaussian Process
  Regression with Application to Safe Control}. In
  \bibinfo{booktitle}{\emph{Advances in Neural Information Processing
  Systems}}, \bibfield{editor}{\bibinfo{person}{H.~Wallach},
  \bibinfo{person}{H.~Larochelle}, \bibinfo{person}{A.~Beygelzimer},
  \bibinfo{person}{F.~d\textquotesingle Alch\'{e}-Buc},
  \bibinfo{person}{E.~Fox}, {and} \bibinfo{person}{R.~Garnett}} (Eds.),
  Vol.~\bibinfo{volume}{32}. \bibinfo{publisher}{Curran Associates, Inc.}
\newblock
\urldef\tempurl%
\url{https://proceedings.neurips.cc/paper_files/paper/2019/file/fe73f687e5bc5280214e0486b273a5f9-Paper.pdf}
\showURL{%
\tempurl}


\bibitem[\protect\citeauthoryear{Lederer, Yang, Jiao, and Hirche}{Lederer
  et~al\mbox{.}}{2023}]%
        {ledererCooperativeControlUncertain2023}
\bibfield{author}{\bibinfo{person}{Armin Lederer}, \bibinfo{person}{Zewen
  Yang}, \bibinfo{person}{Junjie Jiao}, {and} \bibinfo{person}{Sandra Hirche}.}
  \bibinfo{year}{2023}\natexlab{}.
\newblock \showarticletitle{Cooperative Control of Uncertain Multiagent Systems
  via Distributed Gaussian Processes}.
\newblock \bibinfo{journal}{\emph{IEEE Trans. Automat. Control}}
  \bibinfo{volume}{68}, \bibinfo{number}{5} (\bibinfo{year}{2023}),
  \bibinfo{pages}{3091--3098}.
\newblock
\urldef\tempurl%
\url{https://doi.org/10.1109/TAC.2022.3205424}
\showDOI{\tempurl}


\bibitem[\protect\citeauthoryear{Liu, Cai, Wang, and Ong}{Liu
  et~al\mbox{.}}{2018}]%
        {pmlr-v80-liu18a}
\bibfield{author}{\bibinfo{person}{Haitao Liu}, \bibinfo{person}{Jianfei Cai},
  \bibinfo{person}{Yi Wang}, {and} \bibinfo{person}{Yew~Soon Ong}.}
  \bibinfo{year}{2018}\natexlab{}.
\newblock \showarticletitle{Generalized Robust {B}ayesian Committee Machine for
  Large-scale {G}aussian Process Regression}. In
  \bibinfo{booktitle}{\emph{Proceedings of the 35th International Conference on
  Machine Learning}} \emph{(\bibinfo{series}{Proceedings of Machine Learning
  Research}, Vol.~\bibinfo{volume}{80})},
  \bibfield{editor}{\bibinfo{person}{Jennifer Dy} {and}
  \bibinfo{person}{Andreas Krause}} (Eds.). \bibinfo{publisher}{PMLR},
  \bibinfo{pages}{3131--3140}.
\newblock
\urldef\tempurl%
\url{https://proceedings.mlr.press/v80/liu18a.html}
\showURL{%
\tempurl}


\bibitem[\protect\citeauthoryear{Maddalena, Scharnhorst, and Jones}{Maddalena
  et~al\mbox{.}}{2021}]%
        {maddalenaDeterministicErrorBounds2021}
\bibfield{author}{\bibinfo{person}{Emilio~Tanowe Maddalena},
  \bibinfo{person}{Paul Scharnhorst}, {and} \bibinfo{person}{Colin~N. Jones}.}
  \bibinfo{year}{2021}\natexlab{}.
\newblock \showarticletitle{Deterministic error bounds for kernel-based
  learning techniques under bounded noise}.
\newblock \bibinfo{journal}{\emph{Automatica}}  \bibinfo{volume}{134}
  (\bibinfo{year}{2021}), \bibinfo{pages}{109896}.
\newblock
\showISSN{0005-1098}
\urldef\tempurl%
\url{https://doi.org/10.1016/j.automatica.2021.109896}
\showDOI{\tempurl}


\bibitem[\protect\citeauthoryear{Mchutchon and Rasmussen}{Mchutchon and
  Rasmussen}{2011}]%
        {mchutchonGaussianProcessTraining2011}
\bibfield{author}{\bibinfo{person}{Andrew Mchutchon} {and}
  \bibinfo{person}{Carl Rasmussen}.} \bibinfo{year}{2011}\natexlab{}.
\newblock \showarticletitle{Gaussian Process Training with Input Noise}. In
  \bibinfo{booktitle}{\emph{Advances in Neural Information Processing
  Systems}}, \bibfield{editor}{\bibinfo{person}{J.~Shawe-Taylor},
  \bibinfo{person}{R.~Zemel}, \bibinfo{person}{P.~Bartlett},
  \bibinfo{person}{F.~Pereira}, {and} \bibinfo{person}{K.Q. Weinberger}}
  (Eds.), Vol.~\bibinfo{volume}{24}. \bibinfo{publisher}{Curran Associates,
  Inc.}
\newblock
\urldef\tempurl%
\url{https://proceedings.neurips.cc/paper_files/paper/2011/file/a8e864d04c95572d1aece099af852d0a-Paper.pdf}
\showURL{%
\tempurl}


\bibitem[\protect\citeauthoryear{Provost and Hennessy}{Provost and
  Hennessy}{1996}]%
        {provost1996scaling}
\bibfield{author}{\bibinfo{person}{Foster~John Provost} {and}
  \bibinfo{person}{Daniel~N. Hennessy}.} \bibinfo{year}{1996}\natexlab{}.
\newblock \showarticletitle{Scaling up: distributed machine learning with
  cooperation}. In \bibinfo{booktitle}{\emph{Proceedings of the Thirteenth
  National Conference on Artificial Intelligence - Volume 1}} (Portland,
  Oregon) \emph{(\bibinfo{series}{AAAI'96})}. \bibinfo{publisher}{AAAI Press},
  \bibinfo{pages}{74–79}.
\newblock
\showISBNx{026251091X}


\bibitem[\protect\citeauthoryear{Rasmussen and Williams}{Rasmussen and
  Williams}{2006}]%
        {rasmussenGaussianProcessesMachine2006}
\bibfield{author}{\bibinfo{person}{Carl~Edward Rasmussen} {and}
  \bibinfo{person}{Christopher K.~I. Williams}.}
  \bibinfo{year}{2006}\natexlab{}.
\newblock \bibinfo{booktitle}{\emph{Gaussian Processes for Machine Learning}}.
\newblock \bibinfo{publisher}{{MIT Press}}, \bibinfo{address}{{Cambridge,
  Mass}}.
\newblock
\showISBNx{978-0-262-18253-9}
\showLCCN{QA274.4 .R37 2006}


\bibitem[\protect\citeauthoryear{Schürch, Azzimonti, Benavoli, and
  Zaffalon}{Schürch et~al\mbox{.}}{2020}]%
        {schurchRecursiveEstimationSparse2020}
\bibfield{author}{\bibinfo{person}{Manuel Schürch}, \bibinfo{person}{Dario
  Azzimonti}, \bibinfo{person}{Alessio Benavoli}, {and} \bibinfo{person}{Marco
  Zaffalon}.} \bibinfo{year}{2020}\natexlab{}.
\newblock \showarticletitle{Recursive estimation for sparse Gaussian process
  regression}.
\newblock \bibinfo{journal}{\emph{Automatica}}  \bibinfo{volume}{120}
  (\bibinfo{year}{2020}), \bibinfo{pages}{109127}.
\newblock
\showISSN{0005-1098}
\urldef\tempurl%
\url{https://doi.org/10.1016/j.automatica.2020.109127}
\showDOI{\tempurl}


\bibitem[\protect\citeauthoryear{Srinivas, Krause, Kakade, and Seeger}{Srinivas
  et~al\mbox{.}}{2012}]%
        {srinivasInformationTheoreticRegretBounds2012a}
\bibfield{author}{\bibinfo{person}{Niranjan Srinivas}, \bibinfo{person}{Andreas
  Krause}, \bibinfo{person}{Sham~M. Kakade}, {and} \bibinfo{person}{Matthias~W.
  Seeger}.} \bibinfo{year}{2012}\natexlab{}.
\newblock \showarticletitle{Information-Theoretic Regret Bounds for Gaussian
  Process Optimization in the Bandit Setting}.
\newblock \bibinfo{journal}{\emph{IEEE Transactions on Information Theory}}
  \bibinfo{volume}{58}, \bibinfo{number}{5} (\bibinfo{year}{2012}),
  \bibinfo{pages}{3250--3265}.
\newblock
\urldef\tempurl%
\url{https://doi.org/10.1109/TIT.2011.2182033}
\showDOI{\tempurl}


\bibitem[\protect\citeauthoryear{Stirling, Roberts, Zufferey, and
  Floreano}{Stirling et~al\mbox{.}}{2012}]%
        {6224987}
\bibfield{author}{\bibinfo{person}{Timothy Stirling}, \bibinfo{person}{James
  Roberts}, \bibinfo{person}{Jean-Christophe Zufferey}, {and}
  \bibinfo{person}{Dario Floreano}.} \bibinfo{year}{2012}\natexlab{}.
\newblock \showarticletitle{Indoor navigation with a swarm of flying robots}.
  In \bibinfo{booktitle}{\emph{2012 IEEE International Conference on Robotics
  and Automation}}. \bibinfo{pages}{4641--4647}.
\newblock
\urldef\tempurl%
\url{https://doi.org/10.1109/ICRA.2012.6224987}
\showDOI{\tempurl}


\bibitem[\protect\citeauthoryear{Tresp}{Tresp}{2000a}]%
        {10.1162/089976600300014908}
\bibfield{author}{\bibinfo{person}{Volker Tresp}.}
  \bibinfo{year}{2000}\natexlab{a}.
\newblock \showarticletitle{{A Bayesian Committee Machine}}.
\newblock \bibinfo{journal}{\emph{Neural Computation}} \bibinfo{volume}{12},
  \bibinfo{number}{11} (\bibinfo{date}{11} \bibinfo{year}{2000}),
  \bibinfo{pages}{2719--2741}.
\newblock
\showISSN{0899-7667}
\urldef\tempurl%
\url{https://doi.org/10.1162/089976600300014908}
\showDOI{\tempurl}


\bibitem[\protect\citeauthoryear{Tresp}{Tresp}{2000b}]%
        {trespMixturesGaussianProcesses2000}
\bibfield{author}{\bibinfo{person}{Volker Tresp}.}
  \bibinfo{year}{2000}\natexlab{b}.
\newblock \showarticletitle{Mixtures of Gaussian Processes}. In
  \bibinfo{booktitle}{\emph{Advances in Neural Information Processing
  Systems}}, \bibfield{editor}{\bibinfo{person}{T.~Leen},
  \bibinfo{person}{T.~Dietterich}, {and} \bibinfo{person}{V.~Tresp}} (Eds.),
  Vol.~\bibinfo{volume}{13}. \bibinfo{publisher}{MIT Press}.
\newblock
\urldef\tempurl%
\url{https://proceedings.neurips.cc/paper_files/paper/2000/file/9fdb62f932adf55af2c0e09e55861964-Paper.pdf}
\showURL{%
\tempurl}


\bibitem[\protect\citeauthoryear{W.~Wang and Peng}{W.~Wang and Peng}{2017}]%
        {wangCooperativeLearningNeural2017}
\bibfield{author}{\bibinfo{person}{D.~Wang W.~Wang} {and}
  \bibinfo{person}{Z.~H. Peng}.} \bibinfo{year}{2017}\natexlab{}.
\newblock \showarticletitle{Cooperative learning neural network output feedback
  control of uncertain nonlinear multi-agent systems under directed
  topologies}.
\newblock \bibinfo{journal}{\emph{International Journal of Systems Science}}
  \bibinfo{volume}{48}, \bibinfo{number}{12} (\bibinfo{year}{2017}),
  \bibinfo{pages}{2590--2598}.
\newblock
\urldef\tempurl%
\url{https://doi.org/10.1080/00207721.2017.1324923}
\showDOI{\tempurl}


\bibitem[\protect\citeauthoryear{Wang, Yue, Haaland, and Jeff~Wu}{Wang
  et~al\mbox{.}}{2022}]%
        {wang2022gaussian}
\bibfield{author}{\bibinfo{person}{Wenjia Wang}, \bibinfo{person}{Xiaowei Yue},
  \bibinfo{person}{Benjamin Haaland}, {and} \bibinfo{person}{CF Jeff~Wu}.}
  \bibinfo{year}{2022}\natexlab{}.
\newblock \showarticletitle{{Gaussian processes with input location error and
  applications to the composite parts assembly process}}.
\newblock \bibinfo{journal}{\emph{{SIAM/ASA Journal on Uncertainty
  Quantification}}} \bibinfo{volume}{10}, \bibinfo{number}{2}
  (\bibinfo{year}{2022}), \bibinfo{pages}{619--650}.
\newblock


\bibitem[\protect\citeauthoryear{Yan, Yang, Jiang, Teng, Liu, and Wei}{Yan
  et~al\mbox{.}}{2019}]%
        {8867291}
\bibfield{author}{\bibinfo{person}{Zheping Yan}, \bibinfo{person}{Zewen Yang},
  \bibinfo{person}{Anzuo Jiang}, \bibinfo{person}{Yanbin Teng},
  \bibinfo{person}{Xiangling Liu}, {and} \bibinfo{person}{Shilin Wei}.}
  \bibinfo{year}{2019}\natexlab{}.
\newblock \showarticletitle{Coordinated Control for Trajectory Tracking of
  Multiple UUVs with Input Saturation}. In \bibinfo{booktitle}{\emph{OCEANS
  2019 - Marseille}}. \bibinfo{pages}{1--5}.
\newblock
\urldef\tempurl%
\url{https://doi.org/10.1109/OCEANSE.2019.8867291}
\showDOI{\tempurl}


\bibitem[\protect\citeauthoryear{Yan, Yang, Pan, Zhou, and Wu}{Yan
  et~al\mbox{.}}{2020}]%
        {yanVirtualLeaderBased2020}
\bibfield{author}{\bibinfo{person}{Zheping Yan}, \bibinfo{person}{Zewen Yang},
  \bibinfo{person}{Xiaoli Pan}, \bibinfo{person}{Jiajia Zhou}, {and}
  \bibinfo{person}{Di Wu}.} \bibinfo{year}{2020}\natexlab{}.
\newblock \showarticletitle{Virtual leader based path tracking control for
  Multi-UUV considering sampled-data delays and packet losses}.
\newblock \bibinfo{journal}{\emph{Ocean Engineering}}  \bibinfo{volume}{216}
  (\bibinfo{year}{2020}), \bibinfo{pages}{108065}.
\newblock
\showISSN{0029-8018}
\urldef\tempurl%
\url{https://doi.org/10.1016/j.oceaneng.2020.108065}
\showDOI{\tempurl}


\bibitem[\protect\citeauthoryear{Yang, Dai, Dubey, Hirche, and Hattab}{Yang
  et~al\mbox{.}}{2024a}]%
        {yangAAMAS2024}
\bibfield{author}{\bibinfo{person}{Zewen Yang}, \bibinfo{person}{Xiaobing Dai},
  \bibinfo{person}{Akshat Dubey}, \bibinfo{person}{Sandra Hirche}, {and}
  \bibinfo{person}{Georges Hattab}.} \bibinfo{year}{2024}\natexlab{a}.
\newblock \bibinfo{title}{{Pri-GP: Prior-Aware Distributed Gaussian Process
  Regression}}.
\newblock
\newblock


\bibitem[\protect\citeauthoryear{Yang, Dong, Lederer, Dai, Chen, Sosnowski,
  Hattab, and Hirche}{Yang et~al\mbox{.}}{2024b}]%
        {yangCooperativeLearning2024}
\bibfield{author}{\bibinfo{person}{Zewen Yang}, \bibinfo{person}{Songbo Dong},
  \bibinfo{person}{Armin Lederer}, \bibinfo{person}{Xiaobing Dai},
  \bibinfo{person}{Siyu Chen}, \bibinfo{person}{Stefan Sosnowski},
  \bibinfo{person}{Georges Hattab}, {and} \bibinfo{person}{Sandra Hirche}.}
  \bibinfo{year}{2024}\natexlab{b}.
\newblock \bibinfo{title}{Cooperative Learning with Gaussian Processes for
  Euler-Lagrange Systems Tracking Control under Switching Topologies}.
\newblock
\newblock


\bibitem[\protect\citeauthoryear{Yang, Sosnowski, Liu, Jiao, Lederer, and
  Hirche}{Yang et~al\mbox{.}}{2021}]%
        {yangDistributedLearningConsensus2021}
\bibfield{author}{\bibinfo{person}{Zewen Yang}, \bibinfo{person}{Stefan
  Sosnowski}, \bibinfo{person}{Qingchen Liu}, \bibinfo{person}{Junjie Jiao},
  \bibinfo{person}{Armin Lederer}, {and} \bibinfo{person}{Sandra Hirche}.}
  \bibinfo{year}{2021}\natexlab{}.
\newblock \showarticletitle{Distributed Learning Consensus Control for Unknown
  Nonlinear Multi-Agent Systems based on Gaussian Processes}. In
  \bibinfo{booktitle}{\emph{2021 60th IEEE Conference on Decision and Control
  (CDC)}}. \bibinfo{pages}{4406--4411}.
\newblock
\urldef\tempurl%
\url{https://doi.org/10.1109/CDC45484.2021.9683522}
\showDOI{\tempurl}


\bibitem[\protect\citeauthoryear{Yin, Dai, Yang, Shen, Hattab, and Zhao}{Yin
  et~al\mbox{.}}{2023}]%
        {yin2023learningbased}
\bibfield{author}{\bibinfo{person}{Zhenxiao Yin}, \bibinfo{person}{Xiaobing
  Dai}, \bibinfo{person}{Zewen Yang}, \bibinfo{person}{Yang Shen},
  \bibinfo{person}{Georges Hattab}, {and} \bibinfo{person}{Hang Zhao}.}
  \bibinfo{year}{2023}\natexlab{}.
\newblock \showarticletitle{Learning-based Control for PMSM Using Distributed
  Gaussian Processes with Optimal Aggregation Strategy}. In
  \bibinfo{booktitle}{\emph{IECON 2023- 49th Annual Conference of the IEEE
  Industrial Electronics Society}}. \bibinfo{pages}{1--7}.
\newblock
\urldef\tempurl%
\url{https://doi.org/10.1109/IECON51785.2023.10312503}
\showDOI{\tempurl}


\end{thebibliography}


\end{document}